\documentclass[10pt,twocolumn,letterpaper]{article}

\usepackage[pagenumbers]{cvpr} % To force page numbers, e.g. for an arXiv version

\usepackage{pifont}
\usepackage{multirow}
\usepackage{graphicx}
\usepackage{amsfonts}
\usepackage{amssymb}
\usepackage{bbm}
\usepackage{amsmath}
\usepackage[accsupp]{axessibility} 

\usepackage{makecell}
\usepackage{algorithm}
\usepackage{algpseudocode}
\usepackage{siunitx}

\definecolor{cvprblue}{rgb}{0.21,0.49,0.74}
\usepackage[pagebackref,breaklinks,colorlinks,allcolors=cvprblue]{hyperref}

\def\paperID{7238} % *** Enter the Paper ID here
\def\confName{CVPR}
\def\confYear{2025}

\title{SeaLion: Semantic Part-Aware Latent Point Diffusion Models for 3D Generation}

\author{
\hspace{-12pt}
Dekai Zhu$^{1,2,3}$, \quad
Yan Di$^{1}$, \quad
Stefan Gavranovic$^{2}$, \quad and \quad
Slobodan Ilic$^{1,2}$\\ [0.2em]
$^1$ Technical University of Munich\quad
$^2$ Siemens AG \quad
$^3$ Munich Center for Machine Learning \\
\small{\texttt{\{firstname.lastname\}@tum.de}} \\
}

\definecolor{dk}{rgb}{0.0,0.0,0.0}
\newcommand{\dk}[1]{\textcolor{dk}{\textnormal{{#1}}}}

\begin{document}
\maketitle
\begin{abstract}
Denoising diffusion probabilistic models have achieved significant success in point cloud generation, enabling numerous downstream applications, such as generative data augmentation and 3D model editing.
However, little attention has been given to generating point clouds with point-wise segmentation labels, as well as to developing evaluation metrics for this task.
Therefore, in this paper, we present \textbf{SeaLion}, a novel diffusion model designed to generate high-quality and diverse point clouds with fine-grained segmentation labels.
Specifically, we introduce the \textbf{semantic part-aware latent point diffusion} technique, which leverages the intermediate features of the generative models to jointly predict the noise for perturbed latent points and associated part segmentation labels during the denoising process, and subsequently decodes the latent points to point clouds conditioned on part segmentation labels.
To effectively evaluate the quality of generated point clouds, we introduce a novel point cloud pairwise distance calculation method named \textbf{part-aware Chamfer distance} (\textbf{p-CD}).
This method enables existing metrics, such as 1-NNA, to measure both the local structural quality and inter-part coherence of generated point clouds.
Experiments on the large-scale synthetic dataset ShapeNet and real-world medical dataset IntrA, demonstrate that SeaLion achieves remarkable performance in generation quality and diversity, outperforming the existing state-of-the-art model, DiffFacto, by \textbf{13.33\%} and \textbf{6.52\%} on 1-NNA (p-CD) across the two datasets.
Experimental analysis shows that SeaLion can be trained semi-supervised, thereby reducing the demand for labeling efforts.
Lastly, we validate the applicability of SeaLion in generative data augmentation for training segmentation models and the capability of SeaLion to serve as a tool for part-aware 3D shape editing.
Code available at: \url{https://github.com/Dekai21/SeaLion}. 
\end{abstract}    
\section{Introduction}
In the past few years, 3D point cloud generation based on deep neural networks has attracted significant interest and achieved remarkable success in downstream tasks,
such as 2D image to point cloud generation~\citep{fan2017point, jiang2018gal,di2023ccd} and point cloud completion~\citep{yu2021pointr, huang2020pf}.    
However, little effort has been devoted to the generative models capable of generating 3D point clouds with semantic segmentation labels.
Exiting works~\citep{gal2021mrgan, li2022editvae, shu20193d, zhang2024point} can generate point clouds composed of detachable sub-parts.
Nevertheless, these sub-parts lack clear semantic meaning,
hindering the application of generated point clouds in domains such as generative data augmentation for training segmentation models and semantic part-aware 3D shape editing.

\begin{figure}
    \centering
    \includegraphics[width=0.98\linewidth]{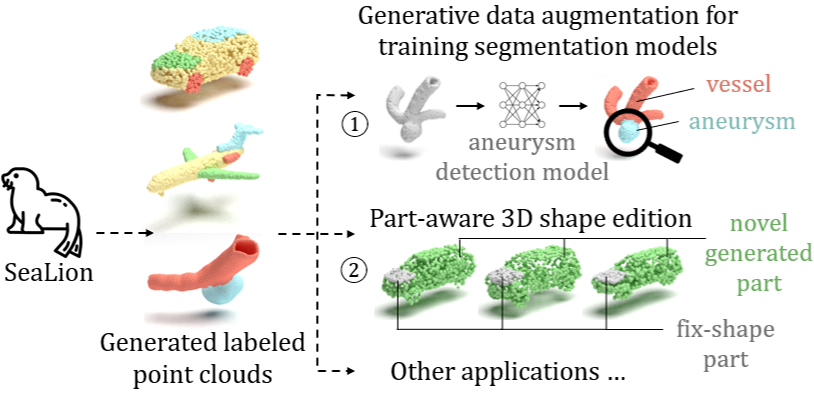}
    \caption{Leveraging the proposed semantic part-aware latent point diffusion technique, \textbf{SeaLion} generates high-quality point clouds with high inter-part coherence and accurate point-wise segmentation labels. The generated data has significant application potential, including enlarging the training sets for data-driven 3D segmentation models, particularly in medical examination domains where labeled data is scarce (\ding{172}). Moreover, SeaLion can serve as \dk{an editing tool, allowing designers to easily replace parts within a 3D shape}. \ding{173} shows examples of generated cars with varying shapes (\textcolor[RGB]{84, 179, 69}{green}) and a fixed-shape hood (\textcolor[RGB]{153, 153, 153}{gray}).}
    \vspace{-1em}
    \label{fig:teaser}
\end{figure}

Attributed to the effective approximation to the real data distribution, denoising diffusion probabilistic models (DDPMs)~\citep{ho2020denoising} outperform many other generative models such as variational autoencoders (VAEs)~\citep{kingma2013auto} and generative adversarial networks (GANs)~\citep{creswell2018generative} in generation quality and diversity.
Current state-of-the-art diffusion-based point cloud generative models~\citep{zeng2022lion, luo2021diffusion, zhou20213d} have achieved impressive performance.
However, they still lack the ability to generate semantic labels.
To the best of our knowledge, DiffFacto~\citep{nakayama2023difffacto} is the only recent work capable of generating point clouds with segmentation labels by utilizing multiple DDPMs to generate each part individually and predicting the pose of each part to assemble the entire point clouds. 
However, due to the part-wise generation factorization, DiffFacto exhibits limited part-to-part coherence within the generated shape.

\noindent \textbf{Semantic part-aware latent point diffusion model.}
Inspired by~\citep{baranchuk2021label}, which demonstrates that the 
intermediate hidden features learned by DDPMs can serve as representations capturing high-level semantic information for downstream vision tasks,
\dk{we propose a novel \textbf{semantic part-aware latent point diffusion} technique.
This technique has two core components: 
(1) It utilizes the latent diffusion model to jointly predict noise for perturbed point-wise latent features and part segmentation labels during the generation process.}
\dk{(2) By incorporating the part segmentation labels as conditional information in the decoder, it enhances the alignment between point coordinates and segmentation labels in the generated point clouds. 
This method yields higher consistency compared to the traditional two-step method, which first generates unlabeled point clouds and subsequently applies a pretrained segmentation model to assign pseudo labels.}
Based on this method, we introduce a generative model named \textbf{SeaLion}. 
Specifically, the point-wise diffusion module in SeaLion includes a down-sampling data path to extract the shared representations for both noise prediction and segmentation tasks, alongside two parallel up-sampling data paths to respectively extract task-specific features. 
\dk{Notably, SeaLion simultaneously diffuses on latent points of all parts, resulting in greater inter-part coherence within a shape compared to DiffFacto.}

\noindent \textbf{Metrics for labeled point cloud generation.}
Commonly used metrics for point cloud generation tasks, such as 1-nearest neighbor accuracy (1-NNA)~\citep{yang2019pointflow} and coverage (COV)~\citep{achlioptas2018learning}, fail to reflect the quality of segmentation-labeled point clouds.
These metrics utilize Chamfer distance (CD) or earth mover's distance (EMD)~\citep{rubner2000earth} to compute the pairwise point cloud distance, neither of which considers the segmentation of point clouds.
\dk{On the other hand, 'ground-truth' segmentation labels are not available for generated samples, making it difficult to use metrics such as mIoU to evaluate label accuracy.}
DiffFacto~\citep{nakayama2023difffacto} assesses each part individually and then averages the results across all parts.
However, this method fails to measure the part-to-part coherence within a shape.
We propose a novel evaluation metric named \textbf{part-aware Chamfer distance} (\textbf{p-CD}) to address these limitations and to
quantify the pairwise distance between two segmentation-labeled point clouds.
Using p-CD, evaluation metrics such as 1-NNA can effectively measure shape plausibility and part-to-part coherence of the generated point clouds.

We conduct extensive experiments on a large-scale synthetic dataset, ShapeNet~\citep{yi2016scalable}, and a real-world 3D intracranial aneurysm dataset, IntrA~\citep{yang2020intra}.
The results show that SeaLion achieves state-of-the-art performance in generating segmentation-labeled point clouds.
Considering that labeling 3D point clouds is tedious, we evaluate SeaLion in a semi-supervised training setting, where only a small portion of the training data is labeled.
Experimental results on ShapeNet validate that SeaLion can leverage additional unlabeled data, highlighting its potential to reduce labeling efforts.
Further studies confirm the feasibility of using point clouds generated by SeaLion for generative data augmentation and \dk{demonstrate SeaLion's potential as an editing tool, allowing designers to easily replace parts within a 3D shape}.
In summary, the contributions of this work are:
\begin{itemize}
    \item We propose a novel generative model named \textbf{SeaLion}, capable of generating high-quality and diverse point clouds with accurate semantic segmentation labels.
    \item We propose a novel distance calculation method named \textbf{part-aware Chamfer distance} (\textbf{p-CD}), enabling metrics such as 1-NNA, COV, and MMD to evaluate the quality and  diversity of segmentation-labeled point clouds.
    \item We demonstrate that SeaLion achieves state-of-the-art performance on a large synthesis dataset, ShapeNet, and a real-world medical dataset, IntrA. Furthermore, we show that SeaLion can be trained in a semi-supervised manner, reducing the need for labeling efforts. 
    \item We confirm the feasibility of generative data augmentation using the point clouds generated by SeaLion and showcase SeaLion's function as an editing tool for part-aware 3D shape editing.
\end{itemize}
\section{Related Works}

\noindent \textbf{Detachable point cloud generation.}
To enable sub-part replacement in point cloud editing, several studies~\citep{shu20193d, li2021sp, gal2021mrgan, li2022editvae, yang2022dsg, hertz2022spaghetti, nakayama2023difffacto} focus on generating point clouds with detachable parts.
TreeGAN~\citep{shu20193d} models point cloud generation as a tree growth process, integrating parts at leaf nodes.
EditVAE~\citep{li2022editvae} learns a disentangled latent representation for each part in an unsupervised manner.
However, these methods do not ensure clear semantic meaning for the parts.
Recently, DiffFacto~\citep{nakayama2023difffacto} addresses this by generating segmented point clouds using multiple DDPMs for each part separately and eventually assemble the entire point clouds. using separate DDPMs for each part and assembling them into a complete shape. To our knowledge, this is the only method generating semantically meaningful parts.

\begin{figure*}[t]
    \centering
    \includegraphics[width=0.95\linewidth]{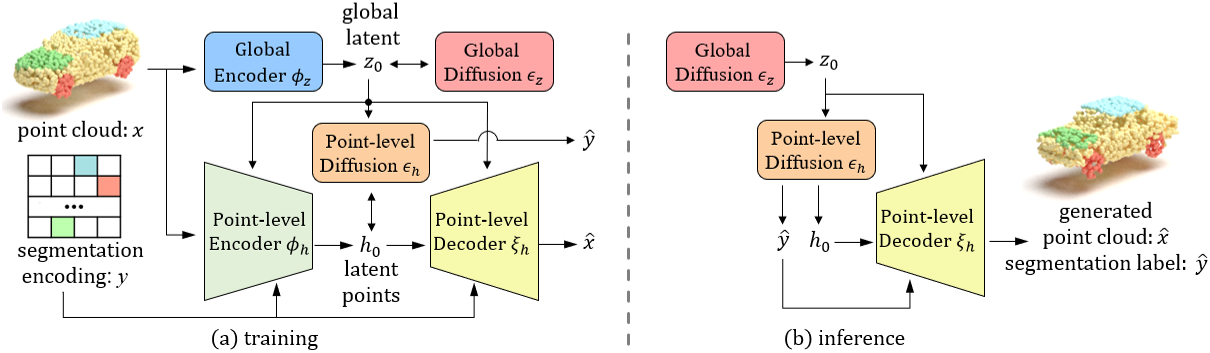}
    \caption{(a) \textbf{Training:} 
    \dk{The generative model develops semantic part awareness by being trained to reconstruct input point clouds $x$ guided by segmentation encodings $y$, and to jointly predict the noise~$\hat{\epsilon}_t$ for perturbed latent points~$h_t$ and segmentation labels~$\hat{y}_t$ at diffusion step~$t$.}
    (b) \textbf{Inference:} Starting from Gaussian noise, the diffusion modules generate $z_0$, $h_0$, and $\hat{y}$. 
    \dk{Then, the conditional decoding guided by $z_0$ and $\hat{y}$ generates a point cloud $\hat{x}$ that maintains strong alignment with $\hat{y}$.}}
    \label{fig:pipeline}
\end{figure*}

\noindent \textbf{Diffusion-based point cloud generation.}
PVD~\citep{zhou20213d} and DPM~\citep{luo2021diffusion} train a diffusion model to generate point clouds directly.
As a state-of-the-art model, Lion~\citep{zeng2022lion} shows that mapping point clouds into regularized latent spaces and training DDPMs to learn the smoothed distributions is more effective than training DDPMs directly on point clouds.

\noindent \textbf{Representations from generative models for discriminative tasks.}
Some recent works explore using generative models as representation learners for discriminative tasks. \citep{donahue2016adversarial, donahue2019large, chen2020generative} use the representations learned by GAN encoders and masked pixel predictors for 2D image classification.
Without any additional training, \citep{li2023your} chooses the category conditioning that best predicts the noise added to the input image as the classification prediction. 
\citep{zhang2021datasetgan, tritrong2021repurposing, xu2021linear, xu2021generative, baranchuk2021label} investigate the usage of generative models on the segmentation tasks.
\citep{baranchuk2021label} shows that intermediate activations capture the semantic information from the input images and appear to be useful representations for the segmentation problem.

\section{Methodology}

In this section, we first give preliminaries on DDPMs~\citep{ho2020denoising} and propose the {semantic part-aware latent points} technique.
Next, we introduce the architecture of {SeaLion}, 
and illustrate its usage as a part-aware 3D edition tool.
Finally, we discuss the limitation of current metrics for evaluating generated labeled point clouds and propose novel metrics based on {part-aware Chamfer distance} (p-CD).

\subsection{Semantic Part-Aware Latent Point Diffusion}
\label{subsec:method:semantic_latent_points}
The diffusion model~\citep{ho2020denoising} generates data by simulating a stochastic $T$-step process.

During training, the diffusion model $\epsilon_{\theta}$ with parameters $\theta$ is trained to predict the noise~$\epsilon$ to denoise the perturbed sample $x_t$ at step $t$. The training loss function is:
\begin{equation}
    \mathcal{L}(\epsilon_{\theta}) = \mathbb{E}_{t, x_0, \epsilon} [{|| \epsilon_{\theta} (x_t, t, a) - \epsilon || }^2_2],
\end{equation}
where $t \sim \text{Uniform}\{1,2,...,T\} $ is the diffusion time step, 
$\epsilon \sim \mathcal{N}(0, I)$ is the noise for diffusing $x_0$ to $x_t$,
and $a$ is the conditional information, such as category encoding.
During inference, the diffusion model starts from a random sample $x_T \sim \mathcal{N}(0, I)$ and denoises it iteratively until $t = 0$.

Given a point cloud $x \in \mathbb{R}^{n \times 3}$ consisting of $n$ points, Lion~\cite{zeng2022lion} maps it to a global latent $z_0 \in \mathbb{R}^{d_z}$ and latent points $h_0 \in \mathbb{R}^{n \times {d_h}}$, and diffuses on these two latent features respectively. 
The $d_z$-dimensional vector $z_0$ encodes the global shape of the point cloud and serves as conditional information for point-level modules,
while latent points $h_0$ encode the point-wise features and preserve the point cloud structure.
However, the lack of semantic awareness of the model hinders the generation of segmentation-labeled data.

\begin{figure}[t]
    \centering
    \includegraphics[width=0.8\linewidth]{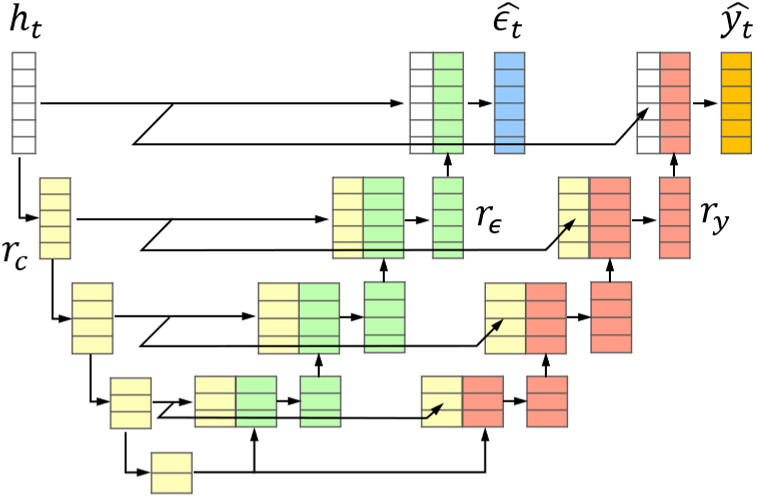}
    \caption{Data flow in the point-level diffusion module $\epsilon_h$. The input, perturbed latent points $h_t$ at step $t$, is down-sampled and transformed to common representations~$r_c$ (\textcolor[RGB]{243, 210, 102}{yellow}). Two parallel up-sampling paths concatenate~$r_c$ with task-specific features, $r_\epsilon$ (\textcolor[RGB]{84, 179, 69}{green}) and $r_y$ (\textcolor[RGB]{240, 67, 50}{red}), to separately predict the noise $\hat{\epsilon_t}$ and the segmentation encoding $\hat{y_t}$.}
    \label{fig:point_diffusion}
\end{figure}

Inspired by the insight that DDPMs can serve as powerful representation learners for discriminative tasks like segmentation~\citep{baranchuk2021label}, we propose \textbf{semantic part-aware latent point diffusion} technique for generating labeled point cloud.
This technique builds on the hierarchical latent diffusion paradigm used in Lion but incorporates segmentation encodings $y \in \mathbb{R}^{n \times c}$ as conditional information for the point-level encoder~$\phi_h: \mathbb{R}^{n \times 3} \times \mathbb{R}^{n \times c} \times \mathbb{R}^{d_z} \rightarrow \mathbb{R}^{n \times d_h}$ and decoder~$\xi_h: \mathbb{R}^{n \times d_h} \times \mathbb{R}^{n \times c} \times \mathbb{R}^{d_z} \rightarrow \mathbb{R}^{n \times 3}$, where $c$ is the number of segmentation parts.
The encoding and decoding processes in the conditional VAE are as follows:
\begin{equation}
    h_0 \leftarrow \phi_h (x, y, z_0), \quad \hat{x} \leftarrow \xi_h (h_0, y, z_0),
\end{equation}
where $\hat{x}$ denotes the reconstructed point cloud that aligns with segmentation encoding~$y$, as illustrated in Figure~\ref{fig:pipeline}~(a).
\dk{The generative model acquires semantic part awareness by being trained to reconstruct input point clouds guided by segmentation encodings, forming a basis for extracting segmentation information from the latent feature $h_0$ in the next step.
To further enhance the generative model's semantic part awareness, it is trained to utilize the intermediate features of the point-level diffusion model~$\epsilon_h: \mathbb{R}^{n \times d_h} \times \mathbb{R} \times \mathbb{R}^{d_z} \rightarrow \mathbb{R}^{n \times d_h} \times \mathbb{R}^{n \times c}$ to jointly predict the noise~$\hat{\epsilon}_t$ for perturbed latent points~$h_t$ and segmentation labels~$\hat{y}_t$} at diffusion step~$t$, as illustrated in Figure~\ref{fig:pipeline}~(a):
\begin{equation}
\hat{\epsilon}_t, \hat{y}_t \leftarrow \epsilon_h (h_t, t, z_0).
\end{equation}
To capture the features at different scales, we utilize a U-Net architecture in~$\epsilon_h$.
Notably, we use a down-sampling data path to extract common representations for both prediction tasks, alongside two parallel up-sampling data paths for extracting task-specific features,  
as illustrated in Figure~\ref{fig:point_diffusion}.
Let $r_c$, $r_\epsilon$, and $r_y$ represent the intermediate features of representation learning, noise prediction, and segmentation prediction, respectively. Given the input $h_t$, the data flow in the down-sampling path is as follows:

\begin{equation}
    r^i_c = 
    \begin{cases}
        {h_t},  &i = 0, \\
        f^i_c (r^{i-1}_c), &i \in \{ 1, ..., U \},
    \end{cases}
\end{equation}
where $f^i_c$ denotes the learnable encoding function at the $i$-th layer of down-sampling path, and $U$ represents the number of layers.
For the noise prediction branch, 
\begin{equation}
    r^i_\epsilon = 
    \begin{cases}
        f^i_\epsilon (r^{i}_c), &i = U , \\
        f^i_\epsilon (r^{i+1}_\epsilon \oplus r^{i}_c), &i \in \{ U\!-\!1, ..., 0 \},
    \end{cases}
\end{equation}
where $f^i_\epsilon$ denotes the learnable encoding function at the $i$-th layer of noise prediction branch, and $\oplus$ is the concatenation operation.
The same paradigm applies to the segmentation prediction branch. 
The final outputs of $\epsilon_h$ are the predicted noise and segmentation labels, i.e. $\hat{\epsilon}_t \leftarrow r^0_\epsilon$ and $\hat{y}_t \leftarrow r^0_y$.

Notably, over the denoising process, $\hat{y}_t$ is progressively smoothed to $\hat{y}$, which serves as conditional information for generating novel point clouds during inference, as illustrated in Figure~\ref{fig:pipeline}~(b). 
\dk{Compared to the traditional two-step method, which first generates unlabeled point clouds and then assigns pseudo segmentation labels using a pretrained segmentation model, our approach is simpler and more robust: (1)~it avoids reliance on an external model; (2) the conditional decoding guided by $\hat{y}$ improves the alignment between $\hat{x}$ and $\hat{y}$, enhancing resilience to mispredictions in $\hat{y}$}.

\noindent \textbf{Training.}
Using this technique, the training consists of two stages.
In the first stage, we train the components of hierarchical VAE, including $\phi_z$,
$\phi_h$,
and $\xi_h$,
to maximize a variational lower bound on the data log-likelihood (ELBO):
\begin{equation}
\label{eq:elbo}
\begin{aligned}
    \mathcal{L}(\phi_z, \phi_h, \xi_h) &= \mathbb{E}_{x, z_0, h_0} \{ \log p_{\xi_h} (x|h_0, y, z_0) \\ 
    &- \lambda_z D_{KL}[q_{\phi_z}(z_0|x)|\mathcal{N}(0, I)] \\
    &- \lambda_h D_{KL} [q_{\phi_h}(h_0|x,y,z_0) | \mathcal{N}(0, I) ] \},
\end{aligned}
\end{equation}
where $q_{\phi_z}$ and $q_{\phi_h}$ are the posterior distribution for sampling $z_0$ and $h_0$, $p_{\xi_h}$ is the prior for reconstruction prediction, and $\lambda_z$ and $\lambda_h$ are the hyperparameters for balancing reconstruction accuracy and Kullback-Leibler regularization.
In the second stage, we train two diffusion modules $\epsilon_z$
and
$\epsilon_h$.
The training objectives for $\epsilon_z$ and $\epsilon_h$ are:
\begin{equation}
\label{eq:latent_ddpm_global}
    \mathcal{L}(\epsilon_z) = \mathbb{E}_{t, z_0, \epsilon} [{|| \epsilon_{z} (z_t, t) - \epsilon || }^2_2],
\end{equation}
and 
\begin{equation}
\label{eq:latent_ddpm_local}
    \mathcal{L}(\epsilon_h) = \mathbb{E}_{t, h_0, \epsilon} [{|| \hat{\epsilon_t} - \epsilon || }^2_2 + \lambda_{seg} H(y, \hat{y_t})],
\end{equation}
where 
$\epsilon \sim \mathcal{N}(0, I)$ denotes the added noise,
$H(\cdot)$ is cross entropy,
and $\lambda_{seg}$ is the hyperparameter for balancing two prediction tasks. 

\noindent \textbf{Inference.}
As illustrated in Figure~\ref{fig:pipeline}~(b), the inference process consists of three steps.
The global diffusion~$\epsilon_z$ firstly generates a global latent~$z_0$.
Conditioning on $z_0$, the point-level diffusion $\epsilon_h$ then generates the latent points $h_0$ and the associated segmentation prediction $\hat{y}$. 
Since $\hat{y_t}$ is predicted at each denoising step, we apply an exponential moving average (EMA) with a smoothing factor of 0.1 to refine $\hat{y_t}$ to $\overline{y}_t$ from step $T$ to 0.
We take $\overline{y}_0$ as the final prediction result $\hat{y}$. 
Lastly, conditioning on $\hat{y}$ and $z_0$, the point-level decoder $\xi_h$ transforms $h_0$ to the generated point cloud $\hat{x}$.

\subsection{Model Architecture of SeaLion}
\label{subsec:method:architecture}
Based on the semantic part-aware latent point diffusion technique, we introduce a novel point cloud generative model named \textbf{SeaLion}.
The architecture of SeaLion is illustrated as follows:

\noindent \textbf{Point-level encoder $\phi_h$ and decoder $\xi_h$.} 
In SeaLion, $\phi_h$ and $\xi_h$ adopt a similar 4-layer Point-Voxel CNN (PVCNN)~\citep{liu2019point} as their backbones.
PVCNN, a U-Net style architecture for point cloud data, uses the set abstraction layer~\citep{Qi2017PointNetpp} and feature propagation layer~\citep{Qi2017PointNetpp} for down-sampling and up-sampling the points. 
Point-voxel convolutions (PVConv) blocks~\citep{liu2019point}, which merge the advantages of point-based and voxel-based methods, are utilized to extract neighboring features at each layer. 
To incorporate the conditional information, the global latent $z_0$ is integrated through the adaptive Group Normalization~\citep{zeng2022lion} in PVConv, 
while the segmentation encoding $y$ is concatenated with the intermediate features at each layer.

\noindent \textbf{Point-level diffusion $\epsilon_h$.} 
As discussed in \ref{subsec:method:semantic_latent_points}, point-level diffusion $\epsilon_h$ contains a down-sampling path to learn the shared representations and two parallel up-sampling paths to extract the task-specific features.
Accordingly, we adopt a modified PVCNN architecture with one down-sampling path and two up-sampling branches. 

\noindent \textbf{Global encoder $\phi_z$ and diffusion $\epsilon_z$.}
We adopt the same architectures as Lion for the two global-level modules. 
The global encoder $\phi_z$ consists of PVConv blocks, set abstraction layers, a max pooling layer, and a multi-layer perceptron. 
The global diffusion $\epsilon_z$ comprises stacked ResNet~\citep{he2016deep}. 
More details regarding the model architecture can be found in the supplementary materials.

\subsection{Part-aware 3D Shape Edition Tool}
\label{subsec:method:tool}

Since SeaLion is semantic part-aware, it can be used as an editing tool for designers to easily replace parts within a 3D shape. 
Given a 3D shape represented by point cloud $x$ consisting of $|P|$ parts, 
where we aim to preserve part $p \in P$ while introducing variations to the remaining parts.
After transforming the point cloud to the latent points $h$, we can freeze the latent points belonging to part $p$ and apply the diffusion-denoise process~\citep{zeng2022lion, meng2022sdedit} on the unfrozen latent points.
In this process, the unfrozen latent points are perturbed for $\tau$ steps ($\tau < T$) and then denoised for the same number of steps.
Due to the stochasticity of the denoising process, the unfrozen latent points will differ after denoising, leading to deformations in the corresponding parts when decoded by $\xi_h$. 
The pseudo code of using SeaLion as an editing tool is provided in the supplementary materials.

\subsection{Evaluation Metrics}
\label{subsec:method:metrics}
\textbf{Notions.}
Given a generated dataset $\mathcal{G} = \{ x^g | x^g \in \mathbb{R}^{n \times 3}\}$ and a real dataset $\mathcal{R} = \{ x^r | x^r \in \mathbb{R}^{n \times 3}\}$, 
both consist of point clouds with $n$ points.
Suppose each point cloud $x \in \mathbb{R}^{n \times 3}$ consists of $|P|$ parts, i.e. $x = \{ x_p | p \in P, x_p \in \mathbb{R}^{n_p \times 3} \} $, where $n_p$ is the number of points in part~$p$.
For example, if $x$ represents a car from ShapeNet~\citep{yi2016scalable}, $P = \{ \text{roof, hood, wheels, body} \} $.

\begin{figure}[t]
    \centering
    \includegraphics[width=0.99\linewidth]{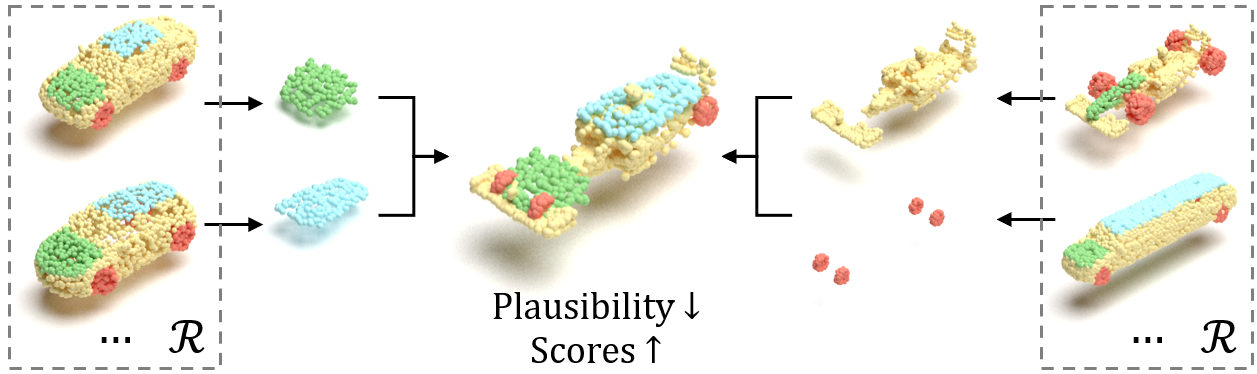}
    \caption{
    Limitations of the intra-part (1-NNA-P) and inter-part (SNAP) scores~\cite{nakayama2023difffacto}. 
    By combining parts from the real dataset $\mathcal{R}$ and maintaining the connection tightness, we can generate a set of implausible samples that still achieves high scores on both metrics.  
    }
    \label{fig:metric}
\end{figure}

\noindent \textbf{Existing metrics.}
The essential of evaluating point cloud generation is to assess both the quality and diversity of the generated data.
Most existing works~\citep{zeng2022lion, zhou20213d, yang2019pointflow} use metrics such as 1-nearest neighbor accuracy~(1-NNA)~\citep{yang2019pointflow}, coverage~(COV), and minimum matching distance~(MMD)~\citep{achlioptas2018learning} for evaluation.
The formulas for these metrics are provided in the supplementary materials.
As discussed in~\citep{yang2019pointflow, zeng2022lion}, COV quantifies generation diversity and is sensitive to mode collapse, but it fails to evaluate the quality of $\mathcal{G}$. 
MMD, on the other hand, only assesses the best quality point clouds in $\mathcal{G}$ and is not a reliable metric to measure overall generation quality and diversity.
1-NNA~\citep{yang2019pointflow} measures both generation quality and diversity by quantifying the distribution similarity between $\mathcal{R}$ and $\mathcal{G}$.
If $\mathcal{G}$ matches $\mathcal{R}$ well, 1-NNA will be close to 50\%.
The aforementioned metrics rely on Chamfer distance (CD) or earth mover's distance (EMD)~\citep{rubner2000earth} to measure the distance between two point clouds.
However, neither CD nor EMD considers the semantic segmentation of the points,
making these metrics ineffective in evaluating the generated point clouds with point-wise segmentation labels.
\dk{Furthermore, the lack of 'ground-truth' segmentation labels on the novel generated data prevents the use of metrics like mIoU for evaluating label accuracy. }
Similar to our work, DiffFacto~\citep{nakayama2023difffacto} tackles this challenge by introducing \textit{intra-part} and \textit{inter-part scores} to evaluate the quality of segmentation-labeled point clouds. 
The intra-part score measures the quality of the independently generated parts and the overall point cloud by averaging the results across all parts.
For example, 1-NNA-P$(\mathcal{R}, \mathcal{G})$~\citep{nakayama2023difffacto} is the average of 1-NNA score for all parts from $\mathcal{R}$ and $\mathcal{G}$, computed by
\begin{equation}
    \frac{1}{|P|} 
    \sum_{p \in P} \frac{ 
 \sum_{x^r_p \in \mathcal{R}_p} \mathbbm{1}[N_{x^r_p} \in \mathcal{R}_p] + \sum_{x^g_p \in \mathcal{G}_p} \mathbbm{1}[N_{x^g_p} \in \mathcal{G}_p] }{|\mathcal{R}_p| + |\mathcal{G}_p|},
\end{equation}
where $\mathcal{G}_p := \{ x^g_p \}$ and $\mathcal{R}_p := \{ x^r_p \}$ represent the generated and real sets of part $p$, respectively, and $\mathbbm{1} [\cdot]$ is the indicator function. $N_{x^r_p}$ is the nearest neighbor of $x^r_p$ in the set $ \mathcal{R}_p \cup \mathcal{G}_p \setminus \{ x^r_p \}$, with the same applying to $N_{x^g_p}$.
The nearest neighbor is determined according to the Chamfer distance. 
Given two parts $x^1_p$ and $x^2_p$, the distance between them, $\text{Chamfer} (x^1_p, x^2_p)$, is computed by
\begin{equation}
    \frac{1}{|x^1_p|} \sum_{q_1 \in x^1_p}{\min_{q_2 \in x^2_p} || q_1 - q_2 ||^2_2} + \frac{1}{|x^2_p|} \sum_{q_2 \in x^2_p}{\min_{q_1 \in x^1_p} || q_1 - q_2 ||^2_2},
\end{equation}
where $q_1, q_2 \in \mathbb{R}^3$ represent points belonging to parts~$x^1_p$ and $x^2_p$, respectively.
The inter-part score, the snapping metric (SNAP)~\citep{nakayama2023difffacto}, measures the connection tightness between two contacting parts in an object. The formula for SNAP is also provided in the supplementary materials. 
However, both intra-part and inter-part scores have limitations in evaluating the generation of segmentation-labeled point clouds.
Specially, averaging the score among all parts or measuring the connection tightness does not effectively measure the coherence among the parts within an object.
An extreme case is illustrated in Figure~\ref{fig:metric}.
By recombining parts from different shapes in the real dataset and maintaining connection tightness, we can create a generated set of implausible samples that still archives high scores on the aforementioned metrics.

\noindent \textbf{Part-aware Chamfer distance.}
To address this issue, we propose part-aware Chamfer Distance (\textbf{p-CD}). Given point clouds $x^1$ and $x^2$ consisting of $P$ parts, the pairwise distance $\text{p-CD}~(x^1, x^2)$ is calculated by

\vspace{-0.8em}
\footnotesize
\begin{equation}
    \sum_{p \in P} \biggl\{ \frac{1}{|x^1_p|} \sum_{q_1 \in x^1_p}{\min_{q_2 \in x^2_p} || q_1 - q_2 ||^2_2} + \frac{1}{|x^2_p|} \sum_{q_2 \in x^2_p}{\min_{q_1 \in x^1_p} || q_1 - q_2 ||^2_2} \biggr\},
\end{equation}
\normalsize
In p-CD, all parts of the point clouds are taken into account.
\dk{For two point clouds consisting of different parts, p-CD is defined as infinite.}
Therefore, if a generated point cloud has a small p-CD to a real point cloud, it indicates that not only are all parts of the generated point cloud of high quality, but they also form a coherent and reasonable assembly as a whole. 
Consequently, the randomly assembled sample in Figure~\ref{fig:metric} will have a large p-CD to the real samples, indicating the anomaly of the generated sample.
Based on p-CD, we can compute the 1-NNA (p-CD), COV (p-CD), and MMD (p-CD) to measure the part-aware proximity of a generated set to a real set.

\section{Experiments}
In this section, we first describe the experimental setup, including the datasets, training details, and evaluation metrics.
Next, we present the evaluation results and the generated point clouds of SeaLion on ShapeNet~\citep{yi2016scalable} and IntrA~\citep{yang2020intra}. 
In the experimental analysis, 
we demonstrate that SeaLion can be trained in a semi-supervised manner, reducing the reliance on labeled data.
Furthermore, we showcase the applicability of SeaLion for generative data augmentation in the point cloud segmentation task and SeaLion's function as a tool for part-aware shape editing.

\subsection{Experimental Setup}

\noindent \textbf{Datasets.}
We conduct extensive experiments on two public datasets, ShapeNet~\citep{yi2016scalable} and IntrA~\citep{yang2020intra}.
ShapeNet~\citep{yi2016scalable} is a large-scale synthetic dataset of 3D shapes with semantic segmentation labels.
We use six categories from this dataset: airplane, car, chair, guitar, lamp, and table.
SeaLion is trained and tested for each category using the official split.
IntrA~\citep{yang2020intra} is a real-world dataset containing 3D intracranial aneurysm point clouds reconstructed from MRI. 
The dataset contains 116 aneurysm segments manually annotated by medical experts. 
We randomly select 93 segments for training and use the remaining 23 segments for testing.
Each aneurysm segment includes the healthy vessel part and the aneurysm part.

\noindent \textbf{Training Details.}
We train SeaLion for each category individually using the training objective functions described in~\ref{subsec:method:semantic_latent_points}. 
The training of SeaLion includes two stages. 
We train the VAE model for 8k epochs in the first stage and the latent diffusion model for 24k epochs in the second stage.
For these two stages, we use an Adam optimizer with a learning rate of 1e-3.
\dk{The parameter sizes of the VAE and diffusion modules in SeaLion are 22.3M and 98.1M, respectively. We conduct the experiments using an NVIDIA RTX 3090 GPU with 24GB of VRAM.}

\noindent \textbf{Metrics.}
We use the part-aware Chamfer distance (p-CD) proposed in \ref{subsec:method:metrics} to quantify the pairwise point cloud distance.
As discussed in~\citep{yang2019pointflow}, 1-NNA measures both generation quality and diversity by computing the distribution similarity between $\mathcal{R}$ and $\mathcal{G}$, while COV and MMD have limitations in measuring the overall generation quality. 
Therefore we compute 1-NNA (p-CD) as the primary evaluation metric in this work, but we still report COV (p-CD), and MMD (p-CD) for convenience of other researchers. 
Additionally, we report the results of 1-NNA-P, COV-P, and MMD-P~\citep{nakayama2023difffacto} for the airplane and chair categories in ShapeNet for comparison to DiffFacto, despite the limitation of these metrics has been illustrated in Section~\ref{subsec:method:metrics} and Figure~\ref{fig:metric}.

\subsection{Experimental Results}

\begin{table*}[t]
\begin{minipage}{.65\linewidth}
\centering
\scalebox{0.84}{
\begin{tabular}{ c @{} c @{} c c c c c c}
\toprule
Metric & Model & Airplane & Car & Chair & Guitar & Lamp & Table \\ \hline

\multirow{4}{*}{\begin{tabular}{@{}c@{}}1-NNA (p-CD) $\downarrow$ \\ (\%) \end{tabular}} & Lion \& PointNet++ & 68.48 & 79.11 & 65.42 & - & - & - \\
& \dk{Lion \& SPoTr} & 67.13 & 77.36 & 65.27 & - & - & - \\
& DiffFacto & 81.67 & 90.51 & 77.34 & - & 67.13 & - \\
& \textbf{SeaLion} & \textbf{65.40} & \textbf{73.10} & \textbf{63.14} & \textbf{62.59} & \textbf{61.71} & \textbf{63.56} \\  \hline

\multirow{4}{*}{\begin{tabular}{@{}c@{}}COV (p-CD) $\uparrow$ \\ (\%) \end{tabular}} & Lion \& PointNet++ & 39.00 & 33.54 & 43.75  & - & - & - \\
& \dk{Lion \& SPoTr} & 42.71 & 35.18 & 44.02 & - & - & -  \\
& DiffFacto & 32.26 & 26.58 & 35.37  & - & 46.95 & - \\
& \textbf{SeaLion} & \textbf{47.51} & \textbf{44.94} & \textbf{46.88} & \textbf{46.85} & \textbf{48.25} & \textbf{41.04} \\  \hline

\multirow{4}{*}{\begin{tabular}{@{}c@{}}MMD (p-CD) $\downarrow$ \\ ($\times 10^{-3}$) \end{tabular}} & Lion \& PointNet++ & \textbf{5.91} & 8.18 & 17.13  & - & - & - \\
& \dk{Lion \& SPoTr} & 6.72 & 8.11 & 16.98 & - & - & - \\
& DiffFacto & 7.15 & 9.03 & 20.30  & - & 29.47 & - \\
& \textbf{SeaLion} & 6.38 & \textbf{7.95} & \textbf{16.25} & \textbf{2.11} & \textbf{28.38} & \textbf{14.56} \\
\bottomrule
\end{tabular}}
\caption{Evaluation on ShapeNet~\citep{yi2016scalable}. Note that certain data is missing because DiffFacto~\citep{nakayama2023difffacto} only provides pretrained models for airplane, car, chair, and lamp categories, while Lion~\citep{zeng2022lion} only releases generated point clouds for airplane, car, and chair categories.}
\label{tab:shapenet}
\end{minipage}
\hfill
\begin{minipage}{0.33\linewidth}
\centering
\includegraphics[width=0.98\linewidth]{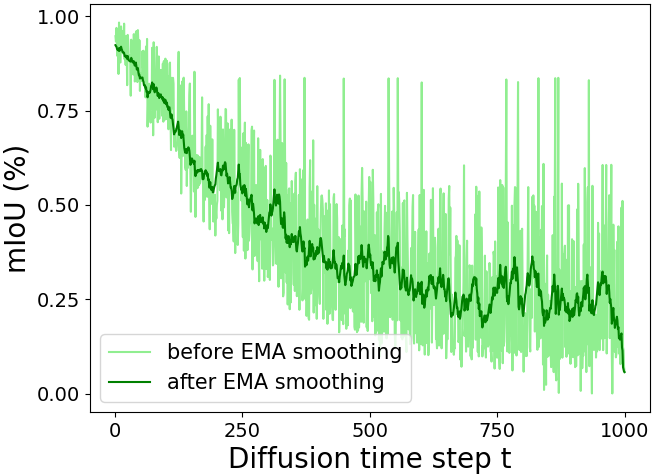}
\captionof{figure}{Evolution of predictive performance measured by mIoU for different diffusion steps $t$ on airplane class. The prediction accuracy improves as $t$ decreases from $T$ to $0$.}
\label{fig:miou}
\end{minipage}
\end{table*}

\begin{figure}[b]
    \centering
    \includegraphics[width=0.98\linewidth]{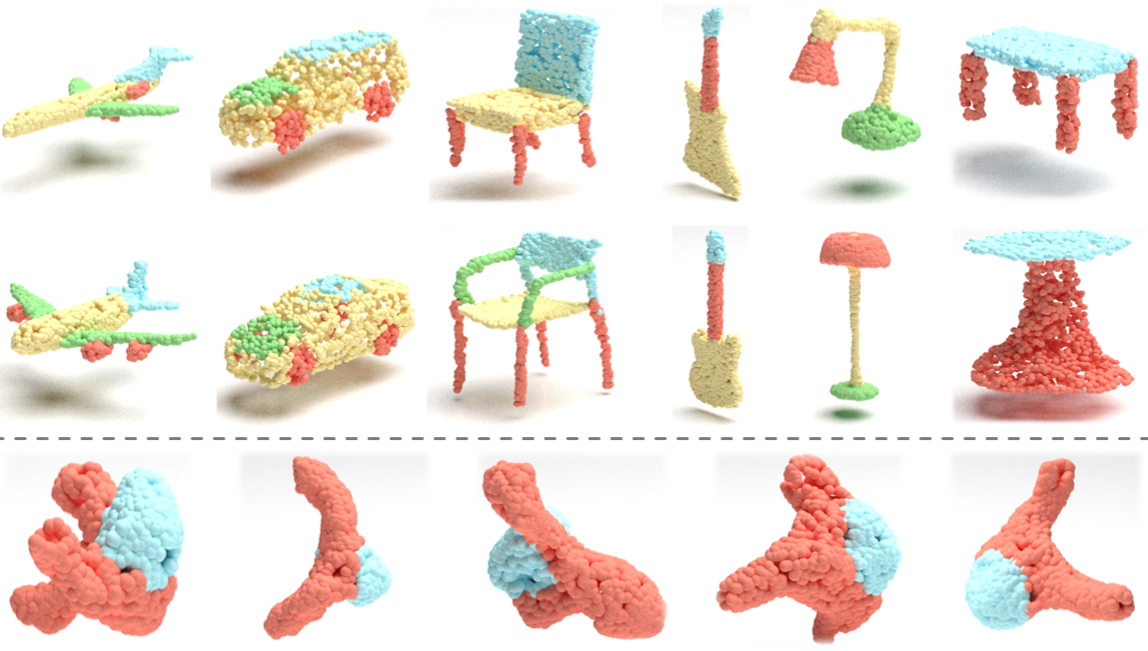}
    \caption{ \textbf{Up:} Generated point clouds of airplanes, cars, chairs, guitars, lamps, and tables from SeaLion. \textbf{Bottom:}~Generated aneurysm segments from SeaLion (\textcolor[RGB]{240, 67, 50}{red}: vessels, \textcolor[RGB]{125, 200, 255}{blue}: aneurysm).}
    \label{fig:shapenet}
\end{figure}

\textbf{Evaluation on ShapeNet.} 
The experimental results of SeaLion on the six classes in ShapeNet are presented in Table~\ref{tab:shapenet}.
DiffFacto~\citep{nakayama2023difffacto} provides pretrained weights for four categories in ShapeNet: airplane, car, chair, and lamp. 
We use these released weights to generate point clouds and evaluate them using our proposed metrics.
Additionally, we use a pretrained PointNet++~\citep{Qi2017PointNetpp} and \dk{SPoTr~\citep{Park2023SelfPositioningPT}, an open-source and state-of-the-art model on ShapeNet part segmentation benchmark~\cite{paperswithcode_shapenet_part},}
to assign pseudo segmentation labels for the officially released point clouds generated from Lion~\citep{zeng2022lion}.
The results demonstrate that SeaLion outperforms both DiffFacto and \dk{the two-step approach, which combines the state-of-the-art generative and segmentation models, Lion and SPoTr.}
For the airplane, car, chair, and lamp categories, SeaLion outperforms DiffFacto by an average of \textbf{13.33\%} on 1-NNA (p-CD), \textbf{11.61\%} on COV (p-CD), and \textbf{10.60\%} on MMD (p-CD), indicating that SeaLion generates higher-quality and more diverse data.
Some of the generated point clouds are demonstrated in Figure~\ref{fig:shapenet}, showing not only plausible shape and part-to-part coherence but also high variety among the shapes.
More generated point clouds are provided in the supplementary materials.
Besides, we report the evaluation of SeaLion according to 1-NNA-P, COV-P, and MMD-P~\citep{nakayama2023difffacto} in Table~\ref{tab:shapenet_old_metrics}. 
The results show that SeaLion outperforms DiffFacto on the primary metric 1-NNA-P and achieves competitive performance on the other metrics.
By comparing the results of DiffFacto~\citep{nakayama2023difffacto} in Table~\ref{tab:shapenet} and Table~\ref{tab:shapenet_old_metrics}, we  observe a notable drop from 1-NNA-P to 1-NNA (p-CD).
\dk{This occurs because 1-NNA (p-CD) measures the implausible inter-part coherence within the shapes generated by DiffFacto. In addition to the extreme case shown in Figure~\ref{fig:metric}, more realistic examples are provided in the supplementary materials.}

In SeaLion, the diffusion~$\epsilon_h$ predicts both noise and segmentation during the generation process.
We demonstrate the evolution of predictive performance, measured by mIoU, across different diffusion steps~$t$ for the airplane category in Figure~\ref{fig:miou}.
As $t$ decreases from $T$ to 0 during the denoising process, the perturbed latent points~$h_t$ become increasingly informative for segmentation prediction.
This trend aligns with the findings in~\citep{baranchuk2021label}.

\noindent \textbf{Evaluation on IntrA.} 
In this experiment, we train SeaLion and DiffFacto~\citep{nakayama2023difffacto} on the IntrA dataset~\citep{yang2020intra} for comparison.
The experimental results presented in Table~\ref{tab:intra} demonstrate that SeaLion outperforms DiffFacto by \textbf{6.52\%} on 1-NNA (p-CD), \textbf{21.74\%} on COV (p-CD), and \textbf{8.45\%} on MMD (p-CD).
Some of the generated intracranial aneurysm segments from SeaLion are presented in Figure~\ref{fig:shapenet}.
These generated samples are of high quality and demonstrate the diverse modalities of aneurysms located in different vessel regions.

\begin{table}
\centering
\scalebox{0.95}{
\begin{tabular}{c  c  c  c }
\toprule
Metric & Model & Airplane & Chair \\ \hline
\multirow{3}{*}{\begin{tabular}{@{}c@{}}1-NNA-P $\downarrow$ \\ (\%) \end{tabular}} & Lion \& PointNet++ & 68.73 & 69.25 \\
& DiffFacto & 68.72 & 65.23\\
& \textbf{SeaLion} & \textbf{68.39} & \textbf{63.24}\\  \hline
\multirow{3}{*}{\begin{tabular}{@{}c@{}}COV-P $\uparrow$ \\ (\%) \end{tabular}} & Lion \& PointNet++ & 38.8 & 35.1 \\
& DiffFacto & \textbf{46.2} & 42.5 \\
& \textbf{SeaLion} & 44.9 & \textbf{46.5}\\  \hline
\multirow{3}{*}{\begin{tabular}{@{}c@{}}MMD-P $\downarrow$ \\ ($\times 10^{-2}$) \end{tabular}} & Lion \& PointNet++ & 3.68 & 3.99\\
& DiffFacto & \textbf{3.20} & 3.27\\
& \textbf{SeaLion} & 3.45 & \textbf{2.73} \\
\bottomrule
\end{tabular}}
\caption{Evaluation of airplane and chair classes in ShapeNet~\citep{yi2016scalable} according to the metrics proposed in DiffFacto~\citep{nakayama2023difffacto}.}
\label{tab:shapenet_old_metrics}
\end{table}

\begin{table}
\centering
\scalebox{0.95}{
\begin{tabular}{c c c }
\toprule
Metric & Model & Aneurysm  \\ \hline
\multirow{2}{*}{1-NNA (p-CD) $\downarrow$ (\%)} & DiffFacto & 71.74 \\
& \textbf{SeaLion} & \textbf{65.22} \\  \hline
\multirow{2}{*}{COV (p-CD) $\uparrow$ (\%)} & DiffFacto & 39.13 \\
& \textbf{SeaLion} & \textbf{60.87} \\  \hline
\multirow{2}{*}{\begin{tabular}{@{}c@{}}MMD (p-CD) $\downarrow$ \\ ($\times 10^{-2}$) \end{tabular}} & DiffFacto & 8.05 \\
& \textbf{SeaLion} & \textbf{7.37} \\
\bottomrule
\end{tabular}}
\caption{Evaluation on IntrA~\citep{yang2020intra}.}
\label{tab:intra}
\end{table}

\subsection{Experimental Analysis}
\label{subsec:exp:ablation}

Compared with collecting 3D data, which can be automated using tools like web crawlers, manually labeling segmentation is tedious and time-consuming.
Therefore, methods for extracting information from unlabeled data have attracted lots of attention in recent years.
Typically, semi-supervised learning effectively reduces the need for extensive data labeling by training models with a combination of a small amount of labeled samples and a larger set of unlabeled samples.
The training process of DiffFacto~\citep{nakayama2023difffacto} involves separate training for each semantic part, which limits its ability to leverage the unsegmented data. 
In contrast, SeaLion generates the points for all parts jointly, making it adaptable to the semi-supervised training approach.
Given an unlabeled sample, we can replace the segmentation encoding~$y$ in \eqref{eq:elbo} with zero padding of the same shape, thereby transforming the corresponding modules to be unconditioned by $y$.
Additionally, we omit the second term $H(y, \hat{y}_t)$ in \eqref{eq:latent_ddpm_local} to skip the training of segmentation prediction on unsegmented samples. 
Consequently, SeaLion can be trained on unlabeled samples using this approach,  while labeled samples can still be processed using the objective functions in~\ref{subsec:method:semantic_latent_points}.
To validate the applicability of SeaLion trained using a semi-supervised approach, we conduct an experiment on the car class in ShapeNet~\citep{yi2016scalable}. 
We randomly select 10\% of the samples in the training set as labeled data, while the remaining 90\% are treated as unsegmented. 
For comparison, we train three models as follows: (1) DiffFacto trained with 10\% labeled data using the official default settings, (2) SeaLion trained with 10\% labeled data, and (3) SeaLion trained in a semi-supervised approach with 10\% labeled data and 90\% unlabeled data. 
The experimental results presented in Table~\ref{tab:ablation:semi} demonstrate that SeaLion outperforms DiffFacto when trained with 10\% labeled data, and its performance further improves after incorporating unlabeled data into the training set.
\dk{Additional ablation studies are provided in the supplementary materials.}

\begin{table}[t]
\centering
\scalebox{0.95}{
\begin{tabular}{c  c  c  c }
\toprule
Metric & Model & Training Set & Car  \\ \hline

\multirow{3}{*}{\begin{tabular}{@{}c@{}}1-NNA (p-CD) $\downarrow$ \\ (\%) \end{tabular}} & DiffFacto  & $\mathcal{L}$ & 90.82 \\
& SeaLion & $\mathcal{L}$ & 87.34 \\
& \textbf{SeaLion} & $\mathcal{L}$ \& $\mathcal{U}$ & \textbf{83.23} \\  \hline

\multirow{3}{*}{\begin{tabular}{@{}c@{}}COV (p-CD) $\uparrow$ \\ (\%) \end{tabular}} & DiffFacto  & $\mathcal{L}$ & 23.42 \\
& SeaLion & $\mathcal{L}$ & 37.34 \\
& \textbf{SeaLion} & $\mathcal{L}$ \& $\mathcal{U}$ & \textbf{41.77} \\  \hline

\multirow{3}{*}{\begin{tabular}{@{}c@{}}MMD (p-CD) $\downarrow$ \\ ($\times 10^{-3}$) \end{tabular}} & DiffFacto  & $\mathcal{L}$ & 9.37 \\
& SeaLion & $\mathcal{L}$ & 8.76 \\
& \textbf{SeaLion} & $\mathcal{L}$ \& $\mathcal{U}$ & \textbf{8.33} \\ 

\bottomrule
\end{tabular}}
\caption{Evaluation of the semi-supervised training on SeaLion. $\mathcal{L}$ refers to the use of 10\% data with segmentation labels, while $\mathcal{U}$ refers to the remaining data without segmentation labels.}
\label{tab:ablation:semi}
\end{table}

\begin{table}[t]
\centering
\scalebox{0.95}{
\begin{tabular}{@{} c @{\hskip 0.2cm} c @{\hskip 0.1cm} c @{\hskip 0.3cm} c @{\hskip 0.3cm} c @{\hskip 0.3cm} c @{\hskip 0.3cm} c @{}}
\toprule
Training Set & Airplane & Car & Chair & Guitar & Lamp & Table \\ \hline
$\mathcal{R}$ & 82.28 & 76.98 & 90.31 & 90.97 & 82.50 & 82.77 \\
\dk{$\mathcal{R}^{\dag}$} & 82.55 & 78.09 & 90.83 & 91.07 & 83.18 & 82.48 \\
$\mathcal{R}^{\dag}$ \& $\mathcal{G}$ & \textbf{83.81} & \textbf{79.43} & \textbf{90.88} & \textbf{91.56} & \textbf{84.54} & \textbf{83.44} \\
\bottomrule
\end{tabular}}
\caption{Generative data augmentation for training SPoTr~\citep{Park2023SelfPositioningPT}. $\dag$~denotes the training set is augmented using traditional methods, which rely on simple geometric transformations.}
\label{tab:gda}
\end{table}

\subsection{Applications}
\noindent \textbf{Generative data augmentation.}
Generative synthetic data has been widely used across various domains to enrich datasets and enhance model performance~\cite{chen2024comprehensive}.
In this experiment, we use the set~$\mathcal{G}$ of point clouds generated by SeaLion to enlarge the real dataset~$\mathcal{R}$ (consisting of fully labeled samples) for training the data-driven segmentation model.
We use SPoTr~\citep{Park2023SelfPositioningPT} to predict the part segmentation across six categories in ShapeNet. 
We evaluate the performance of SPoTr using mIoU.
The results, presented in Table~\ref{tab:gda}, demonstrate that the incorporation of generative data steadily enhances the performance of SPoTr across all categories, \dk{outperforming traditional data augmentation methods that typically rely on simple geometric transformations such as rescaling, rotation, and jittering.}

\begin{figure}[t]
\centering
\includegraphics[width=0.98\linewidth]{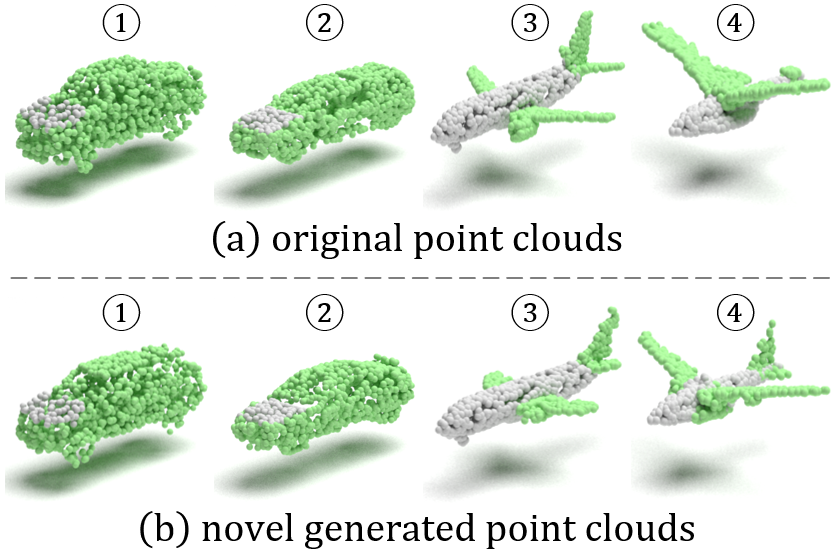}
\captionof{figure}{(a) Original point clouds and (b) novel generated point clouds after part-aware editing (\textcolor[RGB]{153, 153, 153}{gray}: fix-shape parts, \textcolor[RGB]{84, 179, 69}{green}: novel generated parts with deformations).}
\label{fig:part_editing}
\end{figure}

\noindent \textbf{Part-aware 3D shape edition.}
As discussed in \ref{subsec:method:tool}, SeaLion can serve as \dk{an editing tool, allowing designers to easily replace parts within a 3D shape} by running the diffusion-denoise process on the latent points associated with the parts the designers wish to modify.
We conduct experiments on car and airplane point clouds, where the hood of cars and the body of airplanes are selected as the fixed-shape parts.
The experimental results illustrated in Figure~\ref{fig:part_editing} show that novel-generated cars and airplanes keep the chosen parts (hoods and airplane cabins) unchanged while exhibiting diverse deformation in the remaining parts.
\section{Conclusion}
This paper presents a semantic part-aware latent point diffusion technique for generating segmentation-labeled point clouds. 
Using this technique, our model, SeaLion, achieves state-of-the-art performance on ShapeNet and IntrA datasets. 
Additionally, we introduce improved evaluation metrics using a novel part-aware Chamfer distance for evaluating the generated labeled point clouds. 
Extensive experiments demonstrate the effectiveness of SeaLion for generative data augmentation and part-aware 3D shape editing, showcasing its broad applicability in downstream tasks.

{
\clearpage
\small
\bibliographystyle{ieeenat_fullname}
\bibliography{main}
}


% \input{supp}
\appendix
\clearpage
\maketitlesupplementary

\tableofcontents

% \section{Rationale}
% \label{sec:rationale}
% % 
% Having the supplementary compiled together with the main paper means that:
% % 
% \begin{itemize}
% \item The supplementary can back-reference sections of the main paper, for example, we can refer to \cref{sec:intro};
% \item The main paper can forward reference sub-sections within the supplementary explicitly (e.g. referring to a particular experiment); 
% \item When submitted to arXiv, the supplementary will already included at the end of the paper.
% \end{itemize}
% % 
% To split the supplementary pages from the main paper, you can use \href{https://support.apple.com/en-ca/guide/preview/prvw11793/mac#:~:text=Delete%20a%20page%20from%20a,or%20choose%20Edit%20%3E%20Delete).}{Preview (on macOS)}, \href{https://www.adobe.com/acrobat/how-to/delete-pages-from-pdf.html#:~:text=Choose%20%E2%80%9CTools%E2%80%9D%20%3E%20%E2%80%9COrganize,or%20pages%20from%20the%20file.}{Adobe Acrobat} (on all OSs), as well as \href{https://superuser.com/questions/517986/is-it-possible-to-delete-some-pages-of-a-pdf-document}{command line tools}.

\section{Evaluation Metrics}
\subsection{Calculation Formulas}
Given a generated set $\mathcal{G} = \{ x^g | x^g \in \mathbb{R}^{n \times 3}\}$ and a real dataset $\mathcal{R} = \{ x^r | x^r \in \mathbb{R}^{n \times 3}\}$, 
both consist of point clouds with $n$ points.
In practice, $\mathcal{R}$ is the test set unseen during the training of SeaLion, while $\mathcal{G}$ is the set of samples generated during the inference. 
D($\cdot$) is the Chamfer distance or earth mover's distance to measure the distance between two point clouds.
The calculation formulas for metrics such as coverage~(COV), minimum matching distance~(MMD)~\citep{achlioptas2018learning}, 1-nearest neighbor accuracy~(1-NNA)~\citep{yang2019pointflow}, and snapping score (SNAP)~\cite{nakayama2023difffacto} are listed as follows:

\noindent \textbf{Coverage (COV)} measures the ratio of overlap between $\mathcal{R}$ and $\mathcal{G}$ relative to the size of $\mathcal{R}$.
It first constructs a subset by selecting the nearest neighbor in $\mathcal{R}$ for each $x_g$, and then computes the ratio of the cardinality of this subset to the cardinality of $\mathcal{R}$,
% \vspace{-1em}
\begin{equation}
    \text{COV}(\mathcal{G}, \mathcal{R}) = \frac{| \{ \arg\min_{x_r \in \mathcal{R}} D(x_g, x_r ) | x_g \in \mathcal{G} \} |}{|\mathcal{R}|}.
\end{equation}

\noindent \textbf{Minimum matching distance (MMD)} computes the average distance between each $x_r$ in $\mathcal{R}$ and its nearest neighbor in $\mathcal{G}$,
% \vspace{-1em}
\begin{equation}
    \text{MMD}(\mathcal{G}, \mathcal{R}) = \frac{1}{|\mathcal{R}|} \sum_{x_r \in \mathcal{R}} \min_{x_g \in \mathcal{G}} D(x_g, x_r).
\end{equation}

\noindent \textbf{1-nearest neighbor accuracy (1-NNA)} measures the similarity between $\mathcal{R}$ and $\mathcal{G}$ by calculating the proportion of samples in $\mathcal{R}$ or $\mathcal{G}$ whose nearest neighbors belong to the same set.
\small
\begin{equation}
    \text{1-NNA}(\mathcal{G}, \mathcal{R}) = \frac{\sum_{x_g \in \mathcal{G}} \mathbbm{1}(N_{x_g} \in \mathcal{G}) + \sum_{x_r \in \mathcal{R}} \mathbbm{1}(N_{x_r} \in \mathcal{R})}{|\mathcal{G}|+|\mathcal{R}|},
    \label{eq:1-nna-supp}
\end{equation}
\normalsize
where $\mathbbm{1} [\cdot]$ is the indicator function, $N_{x_g}$ is the nearest neighbor of $x_g$ in the set $ \mathcal{R} \cup \mathcal{G} \setminus \{ x_g \}$, with the same applying to $N_{x_r}$.
\dk{\textbf{If $\mathcal{G}$ is very similar to $\mathcal{R}$, it becomes difficult to determine whether the nearest neighbor of $x_g$ belongs to  $\mathcal{G}$ or $\mathcal{R}$}, and vice versa.
In such cases, the 1-NNA score approaches \textbf{50\%}.}

\vspace{0.5em}
\noindent \textbf{Inter-part score (snapping metric, SNAP)}~\citep{nakayama2023difffacto} measures the connection tightness between two contacting parts in a object by computing the Chamfer distance between their closet $N_{\text{SNAP}}$ points, e.g. $N_{\text{SNAP}} = 30$.
For the point cloud $x$, the score SNAP(x) is calculated by
\small
\begin{equation}
    \frac{1}{|P|} \sum_{p_1 \in P} \min_{x_{p_2} \in \mathcal{X}_{p_1} } \text{Chamfer} \{ N^{(N_{\text{SNAP}})}_{x_{p_2}} (x_{p_1}), N^{(N_{\text{SNAP}})}_{x_{p_1}} (x_{p_2}) \},
\end{equation}
\normalsize
where $\mathcal{X}_{p_1}$ denotes the connected parts to $x_{p_1}$, e.g. if $x_{p_1}$ is the car body, $\mathcal{X}_{p_1}$ represents the contacting parts to the car body, \{roof, hood, wheel\}.
$N^{(N_{\text{SNAP}})}_{x_{p_2}} (x_{p_1})$ refers to the $N_{\text{SNAP}}$ nearest points in part $x_{p_1}$ to part $x_{p_2}$.

\subsection{More Discussions about Part-aware Metrics}
In Section~3.4 of the main paper, we introduced novel metrics for evaluating the generation of segmentation-labeled point clouds, including 1-NNA (p-CD).
The formula for 1-NNA is presented at \eqref{eq:1-nna-supp}, while the part-aware Chamfer distance \mbox{p-CD} ($x_1$, $x_2$) between point clouds $x_1$ and $x_2$ is computed as follows:

\footnotesize
\begin{equation}
    \sum_{p \in P} \biggl\{ \frac{1}{|x^1_p|} \sum_{q_1 \in x^1_p}{\min_{q_2 \in x^2_p} || q_1 - q_2 ||^2_2} + \frac{1}{|x^2_p|} \sum_{q_2 \in x^2_p}{\min_{q_1 \in x^1_p} || q_1 - q_2 ||^2_2} \biggr\},
\end{equation}
\normalsize
where $x^p_1$ and $x^p_2$ denote part $p$ of the point clouds $x^1$ and $x^2$, respectively, and $q_1, q_2 \in \mathbb{R}^3$ represent individual points.
For point clouds composed of different parts, we define the pairwise distance as infinity.
Here, we present a more comprehensive discussion on the development and rationale behind 1-NNA (p-CD), as detailed below:
\begin{figure}
    \centering
    \includegraphics[width=0.9\linewidth]{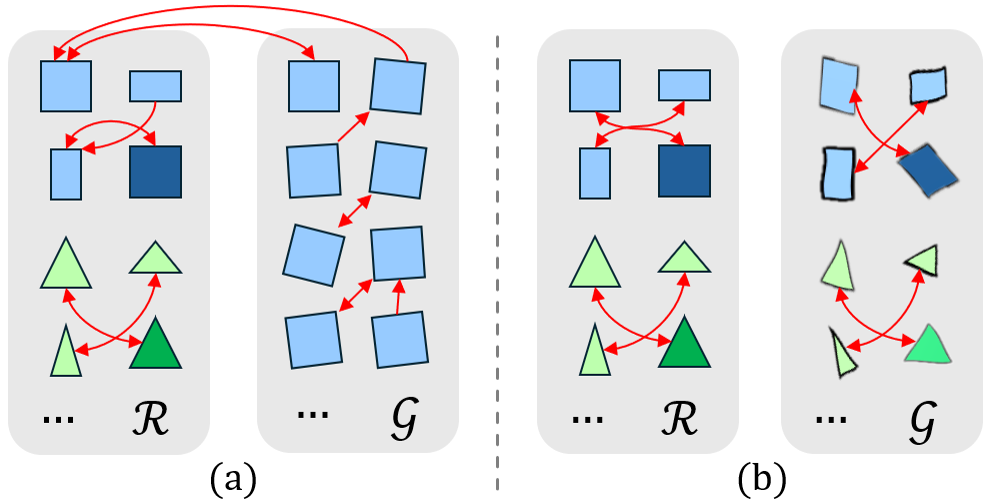}
    \caption{\dk{The generated set $\mathcal{G}$, which either (a) exhibits poor mode coverage compared to the real dataset $\mathcal{R}$ or (b) contains poor-quality samples, cannot achieve a good 1-NNA score. The arrows indicate the nearest neighbors of samples. In both cases, most samples and their nearest neighbors belong to the same set, indicating a significant dissimilarity between $\mathcal{G}$ and $\mathcal{R}$.}}
    \label{fig:1-nna}
\end{figure}
\begin{itemize}
    \item \dk{\textbf{Argument~1}: The core of evaluation for generation tasks is to measure the similarity between the generated set~$\mathcal{G}$ and the real dataset~$\mathcal{R}$. If the two sets cannot be easily distinguished, the performance of the generative model is considered good. The assessment of this distinction incorporates both micro and macro factors: the instance-wise similarity between individual generated and real samples, and the overall distributional similarity between $\mathcal{G}$ and $\mathcal{R}$, i.e. similar mode coverage. In cases where $\mathcal{G}$ consists of high-quality samples but exhibits poor mode coverage (as shown in Figure~\ref{fig:1-nna}~(a)), or where it has similar mode coverage but includes low-quality samples (as shown in Figure~\ref{fig:1-nna}~(b)), the generative model cannot achieve a good 1-NNA score (close to 50\%), since the samples in either $\mathcal{R}$ or $\mathcal{G}$ tend to have their nearest neighbors within the same set.
    \item \textbf{Argument~2}: For the unlabeled generative tasks~\cite{zeng2022lion}, the calculation of 1-NNA is based on Chamfer Distance (CD), which quantifies the shape distance between two point clouds. Thus, the overall quality of $x_g$ is represented by CD($x_g, x_r$): a low value of CD($x_g, x_r$) indicates a ideal quality of $x_g$, and vice versa.
    However, we need to consider two key factors in our task: 
    \subitem i. the overall quality of the generated point clouds, 
    \subitem ii. the accuracy (or rationality) of segmentation. }
    
    \dk{Due to the lack of ``ground truth" segmentation for the generated point clouds, explicitly evaluating segmentation accuracy, such as using mIoU, becomes infeasible.}
    \item \dk{\textbf{Argument 3}: The limitations of metrics such as \mbox{1-NNA-p}~\cite{nakayama2023difffacto}, which obtain final results by averaging part-wise evaluations, in measuring inter-part plausibility are already discussed in Sec.~3.4 of the main paper. In contrast, our novel metric, 1-NNA (p-CD), can \textbf{explicitly evaluate shape quality and implicitly assess the rationality of segmentation}.
    The reasoning is as follows: given $x_g$ and $x_r$ with a very small part-aware Chamfer Distance, i.e. $\text{p-CD}(x_g, x_r)\rightarrow 0$, two facts are implied: 
    \subitem {i.} All parts of $x_g$ are of good quality.
    \subitem {ii.} All parts of $x_g$ align well with the corresponding parts of $x_r$. Since the parts of $x_r$ are assembled in a reasonable way, the corresponding parts of $x_g$ also form a coherent and reasonable whole. In other words, $x_g$ is segmented well. }
    
    \dk{Therefore, 1-NNA (p-CD) effectively measures the similarity of $\mathcal{G}$ and $\mathcal{R}$ from the perspective of overall shape quality and segmentation accuracy.}
\end{itemize}
\section{Pseudo-code of Part-aware 3D Editing}
As discussed in Section 3.3 of the main paper, SeaLion can serve as a tool for part-aware 3D shape editing. The related pseudo code is provided in Algorithm~\ref{alg:pseudo}.

\begin{algorithm*}
\caption{Part-aware 3D shape editing using SeaLion.}
\label{alg:pseudo}
\begin{algorithmic}[1]
\State \textbf{Input:} Point cloud $x$ consisting of $n$ points, segmentation labels $y$, desired fix-shape part $p$.
\State \textbf{Output:} Novel generated point cloud $x_0$ with preserved fix-shape part $p$ and variation in the remaining parts, along with the updated segmentation labels $y_0$.
\State $\text{mask}_p \gets (y == p)$ \Comment{Define a boolean mask to select points belonging to part $p$}
\State $z_0 \gets \phi_z (x)$
\State $h_0 \gets \phi_h (x,y,z_0)$
\State $y_{\tau} \gets y$   \Comment{$\tau < T$}
\State Perturb $h_0$ for $\tau$ steps to $h_{\tau}$
\For{$t \gets \tau $ \textbf{to} $1$}
    % \If{$t == \tau$}
    %     \State $h_t \gets h^{*}_{t}$
    % \EndIf
    \State $h_{t-1}, y_{t-1} \gets \epsilon_h (h_t, t, z_0)$
    \State $y_{t-1} \gets \alpha \cdot y_{t-1} + (1 - \alpha) \cdot y_t$    \Comment{EMA smooth}
    \State $\text{mask}_{p}^{t-1} \gets ((1-\text{mask}_p) \odot y_{t-1} ) == p$
    \State $n_p^{t-1} \gets \sum {\text{mask}_{p}^{t-1}}$
    \If{$ n_p^{t-1} > 0 $}  \Comment{Substitute the latent points in the remaining part but predicted as fix-shape part $p$}
        \State $\text{mask}_\text{others}^{t-1} \gets ((1-\text{mask}_p) \odot y_{t-1} ) != p$
        \State Extract non-zero indices in $\text{mask}_\text{others}^{t-1}$, randomly sample $n_p^{t-1}$ elements and then create a boolean mask for substitution $\text{mask}_\text{resample}^{t-1}$
        \State $h_{t-1}[\text{mask}_{p}^{t-1}] \gets h_{t-1}[\text{mask}_\text{resample}^{t-1}]$
        \State $y_{t-1}[\text{mask}_{p}^{t-1}] \gets y_{t-1}[\text{mask}_\text{resample}^{t-1}]$
    \EndIf
    \State Perturb $h_0$ for $t$ steps to $h^{*}_{t}$
    \State $h_{t-1} \gets \text{mask}_p \odot h^*_{t-1} + (1-\text{mask}_p) \odot h_{t-1}$
    \State $y_{t-1} \gets \text{mask}_p \odot y + (1-\text{mask}_p) \odot y_{t-1}$
\EndFor
\State $x_0 \gets \xi_h (h_0, y_0, z_0)$
\State \textbf{Return} $x_0, y_0$
\end{algorithmic}
\end{algorithm*}
\section{Additional Experimental Details and Results}

\subsection{Two-step Method on IntrA Dataset}
\dk{Since Lion~\cite{zeng2022lion} only released the pretrained weights for airplane, car, and chair classes from ShapeNet~\cite{yi2016scalable}, we retrain Lion on the IntrA~\cite{yang2020intra} dataset to evaluate the two-step method on this dataset. Additionally, we train PointNet++~\citep{Qi2017PointNetpp} and the state-of-the-art segmentation model, SPoTr~\cite{Park2023SelfPositioningPT}, to assign pseudo labels on the generated point clouds, respectively.
The experimental results presented in Table~\ref{tab:intra_supp} demonstrate that SeaLion outperforms DiffFacto and the two-step method across all metrics, aligning with the trends observed in the main paper.}

\begin{table}
\centering
\renewcommand\arraystretch{1.0}
\scalebox{0.95}{
\begin{tabular}{c c c }
\toprule
Metric & Model & Aneurysm  \\ \hline
\multirow{4}{*}{1-NNA (p-CD) $\downarrow$ (\%)} & Lion \& PointNet++ & {74.57} \\
& Lion \& SPoTr & 73.91 \\
& DiffFacto & 71.74 \\
& \textbf{SeaLion} & \textbf{65.22} \\  \hline
\multirow{4}{*}{COV (p-CD) $\uparrow$ (\%)} & Lion \& PointNet++ & {42.65} \\
& Lion \& SPoTr & 30.43 \\
& DiffFacto & 39.13 \\
& \textbf{SeaLion} & \textbf{60.87} \\  \hline
\multirow{4}{*}{\begin{tabular}{@{}c@{}}MMD (p-CD) $\downarrow$ \\ ($\times 10^{-2}$) \end{tabular}} & Lion \& PointNet++ & {8.23} \\
& Lion \& SPoTr & 19.68 \\
& DiffFacto & 8.05 \\
& \textbf{SeaLion} & \textbf{7.37} \\
\bottomrule
\end{tabular}}
\caption{Evaluation on IntrA~\citep{yang2020intra}.}
\label{tab:intra_supp}
\end{table}

\subsection{Impact of the Segmentation Branch}
\begin{table}
\centering
\renewcommand\arraystretch{1.0}
\scalebox{1.00}{
\begin{tabular}{c c c }
\toprule
Metric & Model & Airplane  \\ \hline
\multirow{2}{*}{1-NNA (CD) $\downarrow$ (\%)} & \textbf{Lion} &  \textbf{65.66} \\
& {SeaLion} & {66.27} \\  \hline
\multirow{2}{*}{COV (CD) $\uparrow$ (\%)} & Lion & {46.04} \\
& \textbf{SeaLion} & \textbf{46.63} \\  \hline
\multirow{2}{*}{\begin{tabular}{@{}c@{}}MMD (CD) $\downarrow$ \\ ($\times 10^{-3}$) \end{tabular}} & \textbf{Lion} &  \textbf{3.90} \\
& {SeaLion} & {4.07} \\
\bottomrule
\end{tabular}}
\caption{Impact of SeaLion's segmentation branch on unlabeled generation tasks.}
\label{tab:impact_seg}
\end{table}
\dk{Although DDPMs are increasingly used as representation learners for various downstream tasks~\cite{fuest2024diffusion}, such as image classification and segmentation, we are particularly interested in the impact of the segmentation branch on the original point cloud generation. If the representations for segmentation prediction and 3D noise prediction lie in entirely different distributions, combining both prediction tasks within a unified model could be detrimental.
To investigate this, we ignore the predicted segmentation labels of the point clouds generated by SeaLion and re-evaluate them using metrics designed for unlabeled generative tasks, such as 1-NNA (CD).
It is worth noting that the official weights of Lion~\cite{zeng2022lion} are trained on a larger dataset~\cite{chang2015shapenet} compared to the segmentation-labeled subset~\cite{yi2016scalable}. To ensure a fair comparison, we retrain Lion on the smaller segmentation-labeled subset~\cite{yi2016scalable}.
The experimental results presented in Table~\ref{tab:impact_seg} illustrate that SeaLion achieves performance comparable to Lion in the evaluation of unlabeled generation task. This indicates that the representations for noise and segmentation predictions align well in the feature space. Therefore, incorporating the segmentation prediction branch and its associated training objective do not degrade the generative performance.
}

\subsection{Data Augmentation based on DiffFacto and SeaLion}
Table 5 of the main paper presents the results of generative data augmentation using SeaLion for the segmentation task, where point clouds from six categories generated by SeaLion are incorporated to expand the training set of SPoTr~\citep{Park2023SelfPositioningPT}.
We test on car class using SPoTr with the train set augmented by samples generated by DiffFacto~\citep{nakayama2023difffacto} and SeaLion for comparison.
The results are \textbf{78.23\%} and \textbf{81.43\%} on mIoU.

\subsection{Visualization of Generated Point Clouds from SeaLion}
\begin{figure*}[t]
    \centering
    \includegraphics[width=0.78\linewidth]{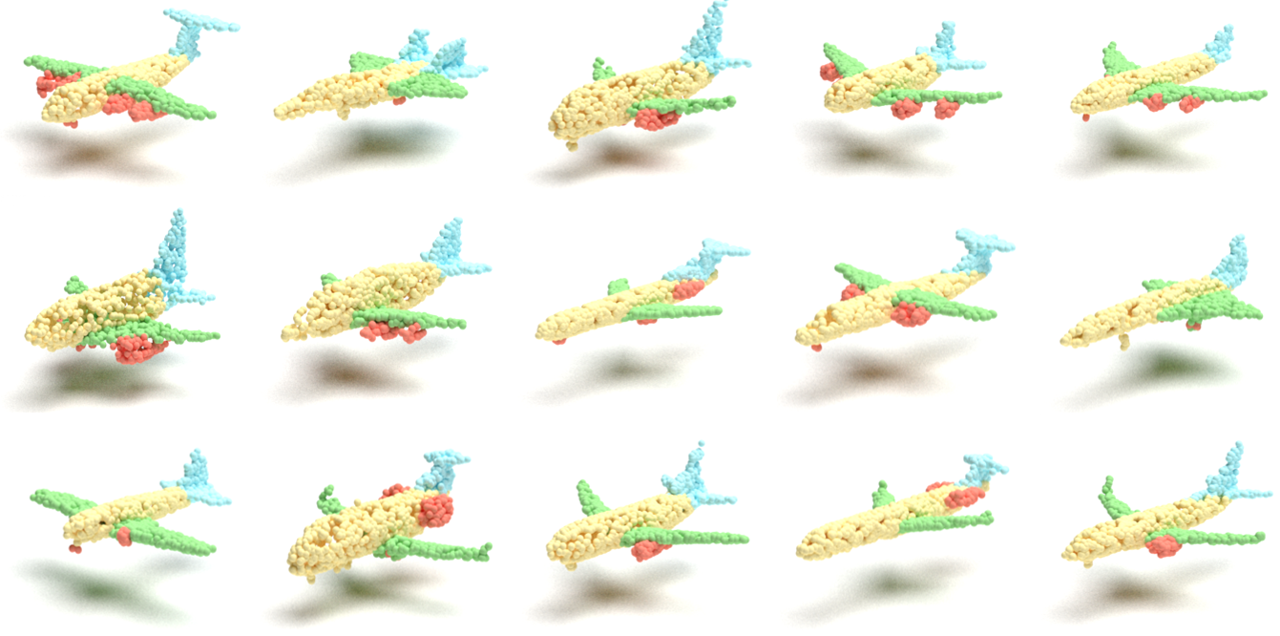}
    \caption{Generated point clouds of airplane class from SeaLion.}
    \label{fig:shapenet_airplane}
\end{figure*}

\begin{figure*}
    \centering
    \includegraphics[width=0.78\linewidth]{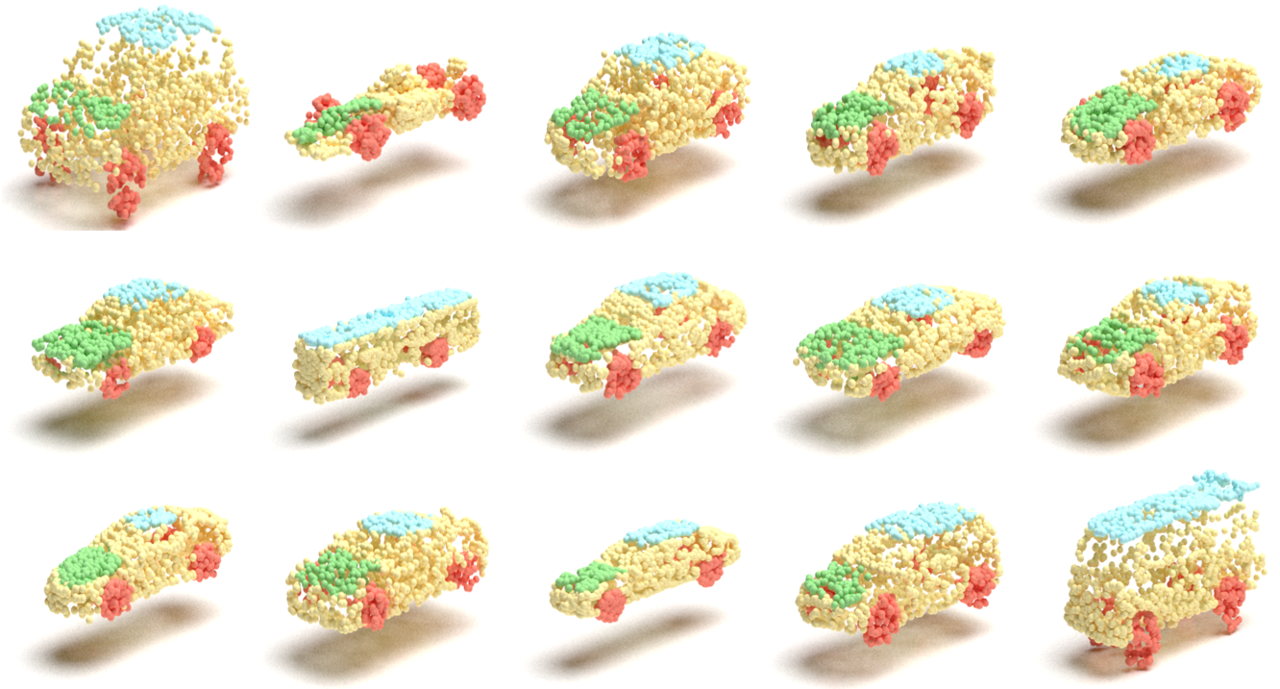}
    \caption{Generated point clouds of car class from SeaLion.}
    \label{fig:shapenet_car}
\end{figure*}

\begin{figure*}
    \centering
    \includegraphics[width=0.78\linewidth]{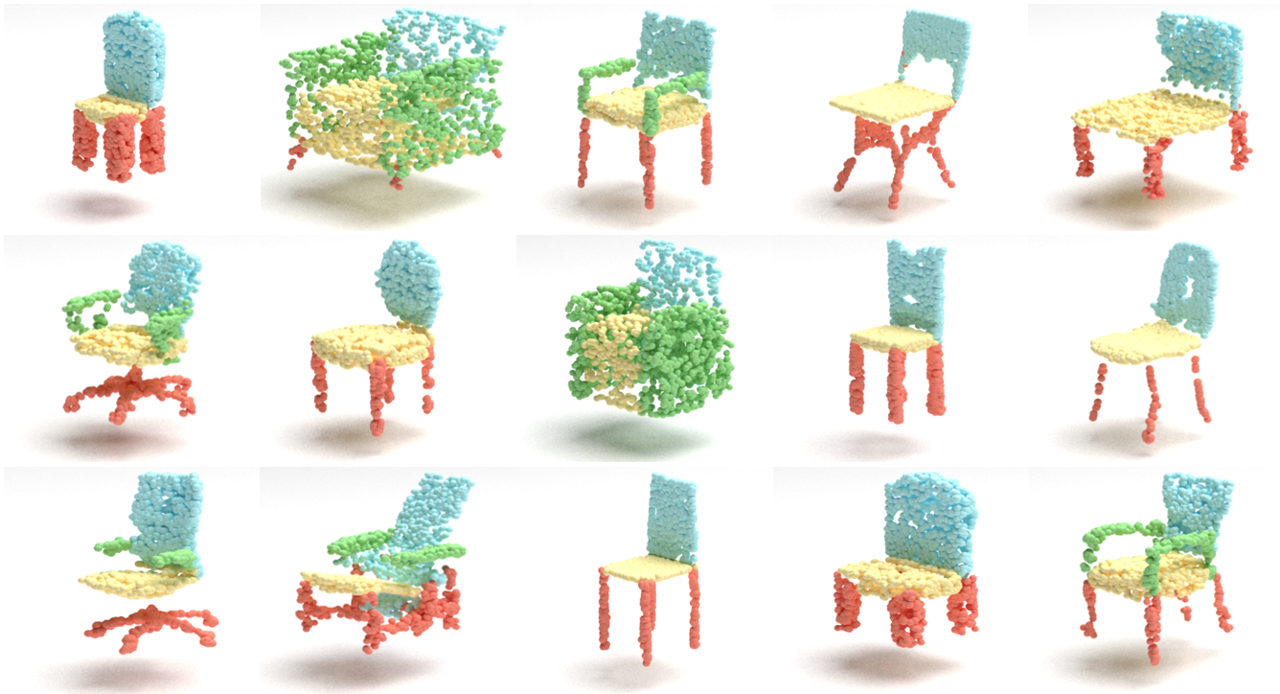}
    \caption{Generated point clouds of chair class from SeaLion.}
    \label{fig:shapenet_chair}
\end{figure*}

\begin{figure*}
    \centering
    \includegraphics[width=0.78\linewidth]{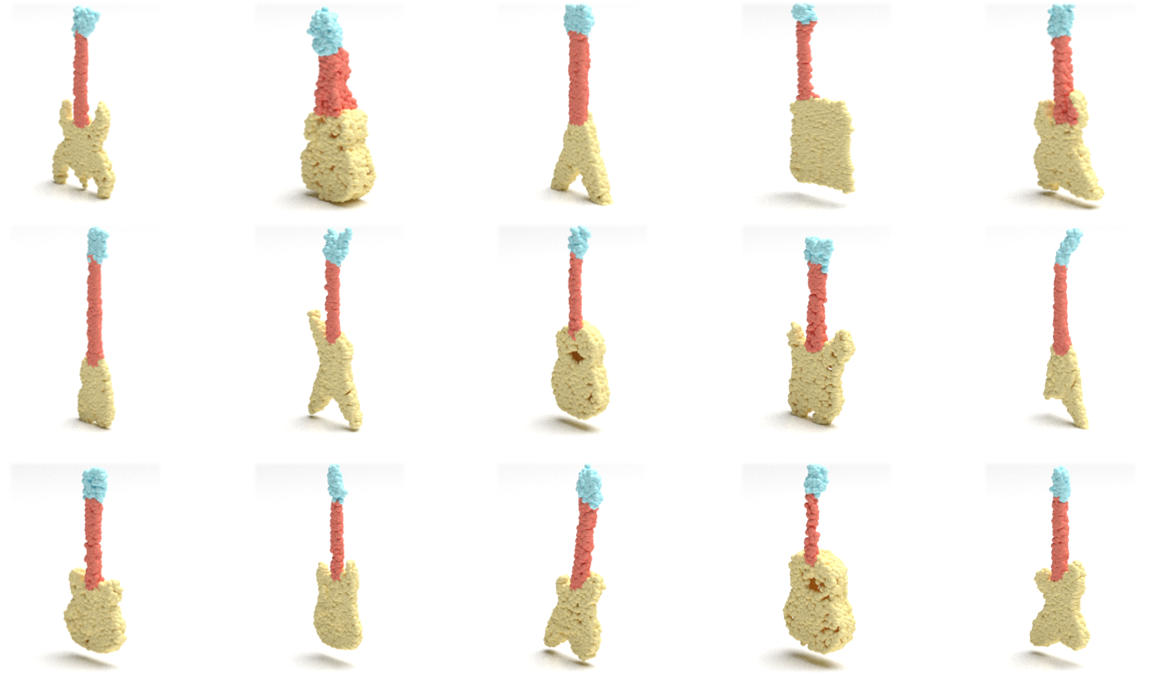}
    \caption{Generated point clouds of guitar class from SeaLion.}
    \label{fig:shapenet_guitar}
\end{figure*}

\begin{figure*}
    \centering
    \includegraphics[width=0.78\linewidth]{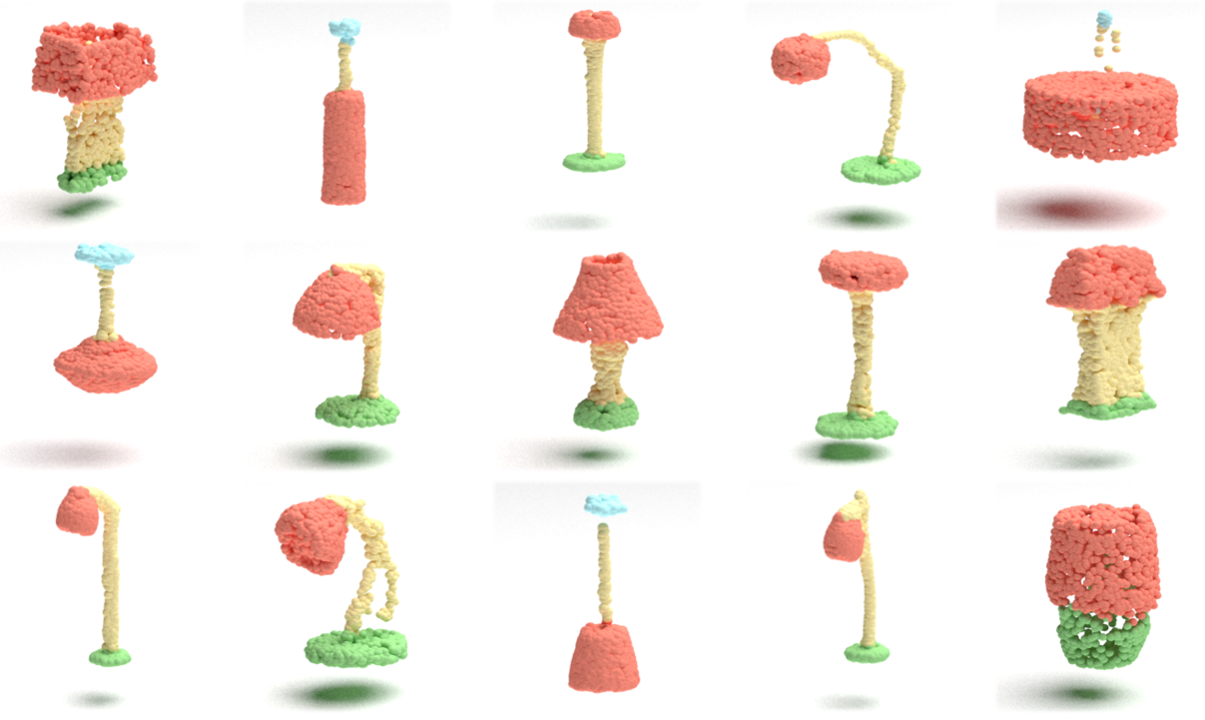}
    \caption{Generated point clouds of lamp class from SeaLion.}
    \label{fig:shapenet_lamp}
\end{figure*}

\begin{figure*}
    \centering
    \includegraphics[width=0.78\linewidth]{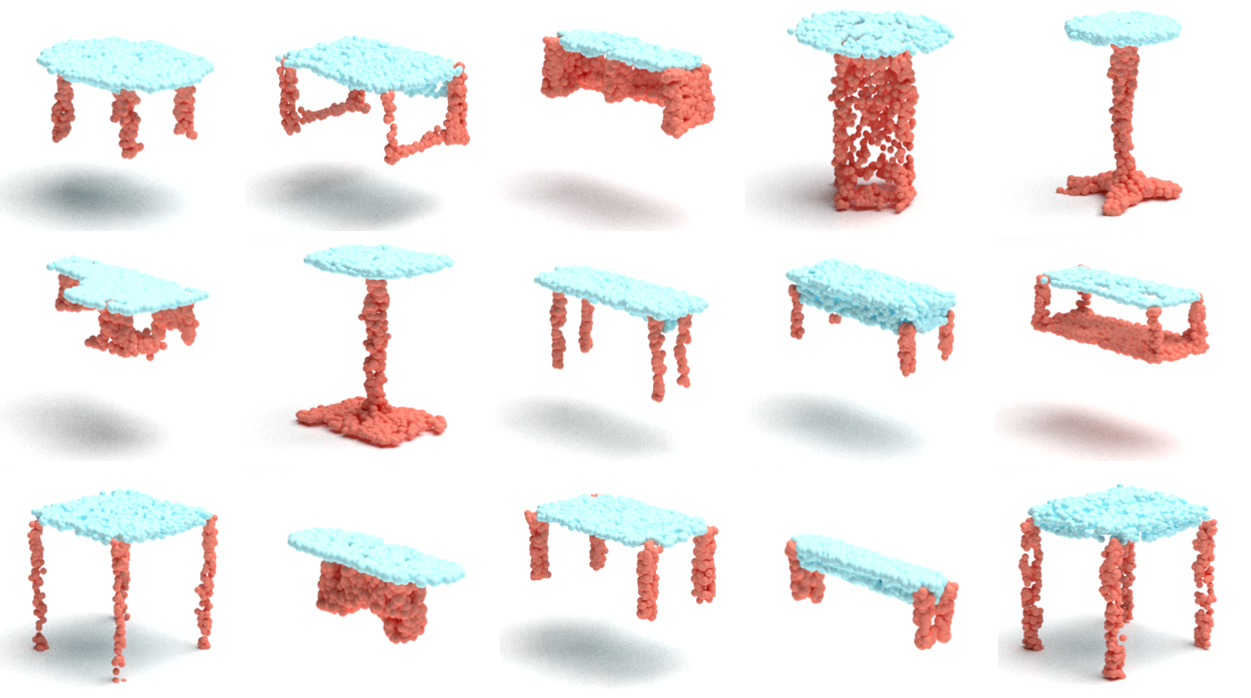}
    \caption{Generated point clouds of table class from SeaLion.}
    \label{fig:shapenet_table}
\end{figure*}

\begin{figure*}
\centering
\includegraphics[width=0.6\linewidth]{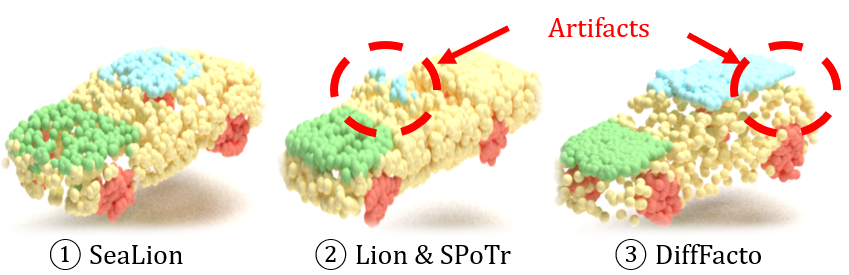}
\caption{Comparison of cars generated by SeaLion, Lion \& SPoTr~\citep{zeng2022lion, Park2023SelfPositioningPT}, and DiffFacto~\citep{nakayama2023difffacto}. In~\ding{173}, the tip of the front grass (yellow) is misclassified as the roof (blue), while in~\ding{174}, the roof part is excessively large and incompatible with the body.}
\label{fig:supp_comparison}
\end{figure*}

Some of the generated point clouds of airplane, car, chair, guitar, lamp, and table categories from SeaLion are demonstrated in Figure~\ref{fig:shapenet_airplane}, \ref{fig:shapenet_car}, \ref{fig:shapenet_chair}, \ref{fig:shapenet_guitar}, \ref{fig:shapenet_lamp}, and \ref{fig:shapenet_table}, respectively.
These generated point clouds demonstrate high-quality on overall shapes and exhibit diverse modalities. 
A video vividly showcasing the point clouds is submitted along with this paper.
Furthermore, Figure~\ref{fig:supp_comparison} presents a visual comparison of cars generated by SeaLion, Lion \& SPoTr~\citep{zeng2022lion, Park2023SelfPositioningPT}, and DiffFacto~\citep{nakayama2023difffacto}.

\subsection{Examples of Implausible Inter-part Coherence within the Generated Point Clouds from \mbox{DiffFacto}}
\begin{figure*}
    \centering
    \includegraphics[width=0.6\linewidth]{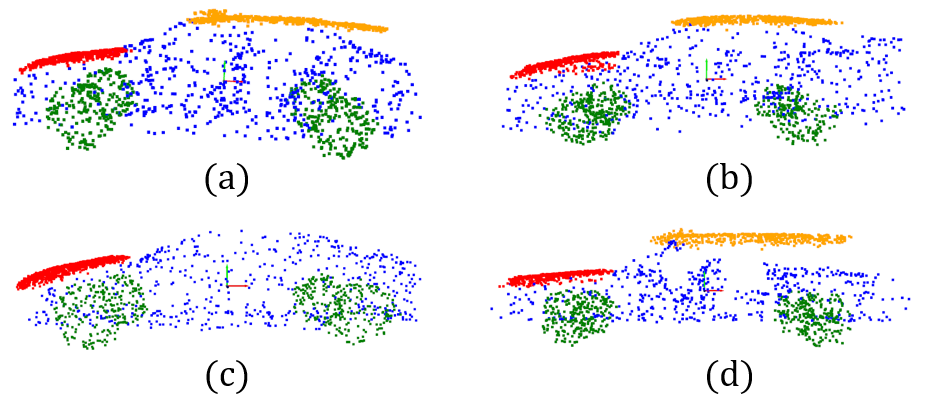}
    \caption{Examples of implausible inter-part coherence in shapes generated by DiffFacto~\cite{nakayama2023difffacto}. (a) \& (b) Too long roof. (c) Hood at an improper position. (d) The convertible car are not supposed to have a flat roof.}
    \label{fig:di3f}
\end{figure*}

\dk{An extreme case of implausible inter-part coherence within a shape is demonstrated in Figure~4 of the main paper. More realistic examples generated from DiffFacto~\cite{nakayama2023difffacto} are shown in Figure~\ref{fig:di3f}.}

\subsection{More Experimental Details and Hyper-parameters}
\textbf{Hyper-parameters of the architecture of SeaLion.}
Details about the hyper-parameters of global encoder $\phi_z$, global diffusion module $\epsilon_z$, point-level encoder $\phi_h$, point-level decoder $\xi_h$, and point-level diffusion module $\epsilon_h$ are listed in Table~\ref{tab:global_encoder}, ~\ref{tab:global_diffusion}, ~\ref{tab:point_encoder}, ~\ref{tab:point_decoder}, and~\ref{tab:point_diffusion}, respectively.
PVConv, SA, GA, and FP refer to point-voxel convolutions modules~\citep{liu2019point}, set abstraction layers~\citep{Qi2017PointNetpp}, global attention layers, and feature propagation layers~\citep{Qi2017PointNetpp}, respectively. 

\noindent \textbf{More training details.}
The training of SeaLion includes two stages. 
We train the VAE model for 8k epochs in the first stage and the latent diffusion model for 24k epochs in the second stage.
For these two stages, we use an Adam optimizer with a learning rate of 1e-3.
We conduct the experiments using an NVIDIA RTX 3090 GPU with 24GB of VRAM.
For the experiments on ShapeNet~\cite{yi2016scalable}, the training process takes an average of 5.4 hours for the first stage and 45 hours for the second stage across six categories.

\noindent \textbf{Details of traditional data augmentation.}
In the experiment of generative data augmentation in the main paper, the traditional data augmentation methods including random rescaling (0.8, 1.2), random transfer (-0.1, 0.1), jittering (-0.005, 0.005), random flipping, random rotation around the x/y/z axis within a small range (-\ang{5}, +\ang{5}).

\begin{table*}
\centering
\renewcommand\arraystretch{1.1}
\begin{tabular}{c|l|c|c}
\toprule
Input & \multicolumn{3}{l}{point clouds ($2048 \times 3$)} \\ \hline
Output & \multicolumn{3}{l}{global latent ($1 \times 128$)} \\ \hline \hline
\multicolumn{2}{l|}{} & Layer 1 & Layer 2 \\ \hline
\multirow{3}{*}{PVConv} & layers & 2 & 1 \\
& hidden dimensions & 32 & 32 \\
& voxel grid size & 32 & 16 \\ \hline
\multirow{5}{*}{SA} & grouper center & 1024 & 256 \\
& grouper radius & 0.1 & 0.2 \\
& grouper neighbors & 32 & 32 \\
& MLP layers & 2 & 2 \\
& MLP output dimensions & 32, 32 & 32, 64 \\ \hline \hline
\multirow{2}{*}{Output layer} & MLP layers & \multicolumn{2}{c}{2} \\
& MLP output dimensions & \multicolumn{2}{c}{128, 128} \\
\bottomrule
\end{tabular}
\caption{Hyper-parameters of the global encoder $\phi_z$.}
\label{tab:global_encoder}
\end{table*}

\begin{table*}
\centering
\renewcommand\arraystretch{1.1}
\begin{tabular}{c|l|c}
\toprule
Input & \multicolumn{2}{l}{global latent ($1 \times 128$), diffusion time step $t$} \\ \hline
Output & \multicolumn{2}{l}{predicted noise on global latent ($1 \times 128$)} \\ \hline \hline
Input linear layer & output dimension & 2048 \\ \hline
\multirow{3}{*}{\makecell[c]{Time embedding\\layer}} & sinusoidal embedding dimension & 128 \\
& MLP layers & 2 \\
& MLP output dimensions & 512, 2048 \\ \hline
\multirow{4}{*}{Stacked ResNet} & MLP layers & 2 \\
& MLP output dimensions & 2048, 2048 \\
& SE MLP layers & 2 \\
& SE MLP output dimensions & 256, 2048 \\ \hline
Output linear layer & output dimension & 128 \\
\bottomrule
\end{tabular}
\caption{Hyper-parameters of the global diffusion $\epsilon_z$.}
\label{tab:global_diffusion}
\end{table*}

\begin{table*}
\centering
\renewcommand\arraystretch{1.1}
\begin{tabular}{c|l|c|c|c|c}
\toprule
Input & \multicolumn{5}{l}{\makecell[l]{point clouds ($2048\times3$), segmentation labels ($2048 \times c$), \\global latent ($1 \times 128$)}} \\ \hline
Output & \multicolumn{5}{l}{point-level latent ($2048 \times 4$)} \\ \hline \hline 
\multicolumn{2}{l|}{} & Layer 1 & Layer 2 & Layer 3 & Layer 4  \\ \hline
\multirow{3}{*}{PVConv} & layers & 2 & 1 & 1 & - \\
& hidden dimensions & 32 & 64 & 128 & - \\
& voxel grid size & 32 & 16 & 8 & - \\ \hline
\multirow{5}{*}{SA} & grouper center & 1024 & 256 & 64 & 16 \\
& grouper radius & 0.1 & 0.2 & 0.4 & 0.8 \\
& grouper neighbors & 32 & 32 & 32 & 32 \\
& MLP layers & 2 & 2 & 2 & 3 \\
& MLP output dimensions & 32, 32 & 64, 128 & 128, 256 & 128, 128, 128 \\ \hline
\multirow{2}{*}{GA} & hidden dimensions & 32 & 128 & 256 & 128 \\
& attention heads & 8 & 8 & 8 & 8 \\ \hline \hline
\multirow{2}{*}{FP} & MLP layers & 3 & 2 & 2 & 2 \\
& MLP output dimensions & 128, 128, 64 & 128, 128 & 128, 128 & 128, 128 \\ \hline
\multirow{3}{*}{PVConv} & layers & 2 & 2 & 3 & 3 \\
& hidden dimensions & 64 & 128 & 128 & 128 \\
& voxel grid size & 32 & 16 & 8 & 8 \\
\bottomrule
\end{tabular}
\caption{Hyper-parameters of the point-level encoder $\phi_h$. Note: layer 1 refers to the shallowest layer and layer 4 refers to the deepest layer, $c$ denotes the number of parts.}
\label{tab:point_encoder}
\end{table*}

\begin{table*}
\centering
\renewcommand\arraystretch{1.0}
\begin{tabular}{c|l|c|c|c|c}
\toprule
Input & \multicolumn{5}{l}{\makecell[l]{point-level latent ($2048 \times 4$), segmentation labels ($2048 \times c$), \\ global latent ($1 \times 128$) } }\\ \hline
Output & \multicolumn{5}{l}{point cloud ($2048 \times 3$)} \\ \hline \hline 
\multicolumn{2}{l|}{} & Layer 1 & Layer 2 & Layer 3 & Layer 4  \\ \hline
\multirow{3}{*}{PVConv} & layers & 2 & 1 & 1 & - \\
& hidden dimensions & 32 & 64 & 128 & - \\
& voxel grid size & 32 & 16 & 8 & - \\ \hline
\multirow{5}{*}{SA} & grouper center & 1024 & 256 & 64 & 16 \\
& grouper radius & 0.1 & 0.2 & 0.4 & 0.8 \\
& grouper neighbors & 32 & 32 & 32 & 32 \\
& MLP layers & 2 & 2 & 2 & 3 \\
& MLP output dimensions & 32, 64 & 64, 128 & 128, 256 & 128, 128, 128 \\ \hline
\multirow{2}{*}{GA} & hidden dimensions & 64+c & 128+c & 256+c & 128+c \\
& attention heads & 8 & 8 & 8 & 8 \\ \hline \hline
\multirow{2}{*}{FP} & MLP layers & 3 & 2 & 2 & 2 \\
& MLP output dimensions & 128, 128, 64 & 128, 128 & 128, 128 & 128, 128 \\ \hline
\multirow{3}{*}{PVConv} & layers & 2 & 2 & 3 & 3 \\
& hidden dimensions & 64 & 128 & 128 & 128 \\
& voxel grid size & 32 & 16 & 8 & 8 \\ \hline \hline
\multirow{2}{*}{Output layer} & MLP layers & \multicolumn{4}{c}{2} \\
& MLP output dimensions & \multicolumn{4}{c}{128, 3} \\
\bottomrule
\end{tabular}
\caption{Hyper-parameters of the point-level decoder $\xi_h$. Note: layer 1 refers to the shallowest layer and layer 4 refers to the deepest layer, $c$ denotes the number of parts.}
\label{tab:point_decoder}
\end{table*}

\begin{table*}
\centering
\renewcommand\arraystretch{1.1}
\begin{tabular}{@{}c|l|c|c|c|c@{}}
\toprule
Input & \multicolumn{5}{l}{\makecell[l]{point-level latent ($2048 \times 4$), diffusion time step $t$, \\ global latent ($1 \times 128$) } }\\ \hline
Output & \multicolumn{5}{l}{\makecell[l]{predicted noise on point-level latent ($2048 \times 4$), \\ predicted segmentation labels ($2048 \times c$) } }\\ \hline \hline 
\multirow{3}{*}{\makecell[c]{Time\\embedding}} & sinusoidal dimensions & \multicolumn{4}{c}{64} \\
& MLP layers & \multicolumn{4}{c}{2} \\
& MLP output dimensions & \multicolumn{4}{c}{64, 64} \\ \hline \hline
\multicolumn{2}{l|}{} & Layer 1 & Layer 2 & Layer 3 & Layer 4  \\ \hline
\multirow{3}{*}{PVConv} & layers & 2 & 1 & 1 & - \\
& hidden dimensions & 32 & 64 & 128 & - \\
& voxel grid size & 32 & 16 & 8 & - \\ \hline
\multirow{5}{*}{SA} & grouper center & 1024 & 256 & 64 & 16 \\
& grouper radius & 0.1 & 0.2 & 0.4 & 0.8 \\
& grouper neighbors & 32 & 32 & 32 & 32 \\
& MLP layers & 2 & 2 & 2 & 3 \\
& MLP output dimensions & 32, 64 & 64, 128 & 128, 256 & 128, 128, 128 \\ \hline
\multirow{2}{*}{GA} & hidden dimensions & 64 & 128 & 256 & 128 \\
& attention heads & 8 & 8 & 8 & 8 \\ \hline \hline
\multirow{2}{*}{\makecell[c]{FP\\(noise)}} & MLP layers & 3 & 2 & 2 & 2 \\
& MLP output dimensions & 128, 128, 64 & 128, 128 & 128, 128 & 128, 128 \\ \hline
\multirow{3}{*}{\makecell[c]{PVConv\\(noise)}} & layers & 2 & 2 & 3 & 3 \\
& hidden dimensions & 64 & 128 & 128 & 128 \\
& voxel grid size & 32 & 16 & 8 & 8 \\ \hline
\multirow{2}{*}{\makecell[c]{Output layer\\(noise)}} & MLP layers & \multicolumn{4}{c}{2} \\
& MLP output dimensions & \multicolumn{4}{c}{128, 4} \\ \hline \hline

\multirow{2}{*}{\makecell[c]{FP\\(segmentation)}} & MLP layers & 3 & 2 & 2 & 2 \\
& MLP output dimensions & 128, 128, 64 & 128, 128 & 128, 128 & 128, 128 \\ \hline
\multirow{3}{*}{\makecell[c]{PVConv\\(segmentation)}} & layers & 2 & 2 & 3 & 3 \\
& hidden dimensions & 64 & 128 & 128 & 128 \\
& voxel grid size & 32 & 16 & 8 & 8 \\ \hline
\multirow{2}{*}{\makecell[c]{Output layer\\(segmentation)}} & MLP layers & \multicolumn{4}{c}{2} \\
& MLP output dimensions & \multicolumn{4}{c}{128, c} \\

\bottomrule
\end{tabular}
\caption{Hyper-parameters of the point-level diffusion $\epsilon_h$. Note: layer 1 refers to the shallowest layer and layer 4 refers to the deepest layer, $c$ denotes the number of parts.}
\label{tab:point_diffusion}
\end{table*}

\end{document}